\pdfoutput=1
\documentclass{article} % For LaTeX2e
\usepackage{iclr2022_conference,times}

\usepackage{amsmath,amsfonts,bm}

\def\eqref#1{equation~\ref{#1}}
\def\1{\bm{1}}

\DeclareMathAlphabet{\mathsfit}{\encodingdefault}{\sfdefault}{m}{sl}
\SetMathAlphabet{\mathsfit}{bold}{\encodingdefault}{\sfdefault}{bx}{n}

\def\gE{{\mathcal{E}}}

\def\gG{{\mathcal{G}}}

\def\gN{{\mathcal{N}}}

\def\gV{{\mathcal{V}}}

\DeclareMathOperator*{\argmin}{arg\,min}

\usepackage[utf8]{inputenc} % allow utf-8 input
\usepackage[T1]{fontenc}    % use 8-bit T1 fonts
\usepackage{hyperref}       % hyperlinks
\usepackage{url}            % simple URL typesetting
\usepackage{booktabs}       % professional-quality tables
\usepackage{amsfonts}       % blackboard math symbols
\usepackage{nicefrac}       % compact symbols for 1/2, etc.
\usepackage{microtype}      % microtypography

\usepackage{amsthm, amssymb, amsmath}
\usepackage{verbatim, float}
\usepackage{mathrsfs}
\usepackage{graphics}
\usepackage{subfig}
\usepackage[usenames,dvipsnames]{pstricks}
\usepackage{multirow}
\usepackage{url}
\usepackage{array}
\usepackage{tabu}
\usepackage{epsfig}
\usepackage{parallel}
\usepackage{graphicx}
\usepackage{diagbox}
\usepackage{mathtools}
\usepackage[absolute,overlay]{textpos}
\usepackage{wrapfig}
\usepackage[linesnumbered,vlined,ruled,commentsnumbered]{algorithm2e}
\newcommand{\proj}{\text{PEG}\xspace}

\theoremstyle{definition}
\newtheorem{theorem}{Theorem}[section]

\newtheorem{definition}[theorem]{Definition}
\newtheorem{lemma}[theorem]{Lemma}

\theoremstyle{remark}

\newcommand{\hide}[1]{}

\title{Equivariant and Stable Positional Encoding for More Powerful Graph Neural Networks\vspace{-2.5mm} }

\author{Haorui Wang$^{1}$\thanks{Wang was an intern at Purdue University when doing this project.} , Haoteng Yin$^2$, Muhan Zhang$^{3,4}$, Pan Li$^{2}$ \\
$^1$School of Computer Science, Wuhan University \\
$^2$Department of Computer Science, Purdue University \\
$^3$Institute for Artificial Intelligence, Peking University, \& $^4$BIGAI \\ % Beijing Institute for General Artificial Intelligence\\ %\\
\texttt{hr\_wang@whu.edu.cn},\,\texttt{muhan@pku.edu.cn},\,\texttt{\{yinht,panli\}@purdue.edu}}
\iclrfinalcopy % Uncomment for camera-ready version, but NOT for submission.
\begin{document}

\maketitle
\vspace{-5mm}
\begin{abstract}
\vspace{-3.5mm}
%\pan{Todo: abstract}
%Many graph-based learning tasks are to make prediction over a set of nodes such as link prediction. 
Graph neural networks (GNN) have shown great advantages in many graph-based learning tasks but often fail to predict accurately for a task based on sets of nodes such as link/motif prediction and so on. Many works have recently proposed to address this problem by using random node features or node distance features. However, they suffer from either slow convergence, inaccurate prediction or high complexity. In this work, we revisit GNNs that allow using positional features of nodes given by positional encoding (PE) techniques such as Laplacian Eigenmap, Deepwalk, etc.. GNNs with PE often get criticized because  they are not generalizable to unseen graphs (inductive) or stable. Here, we study these issues in a principled way and propose a provable solution, a class of GNN layers termed \proj with rigorous mathematical analysis. \proj uses separate channels to update the original node features and positional features. \proj imposes permutation equivariance w.r.t. the original node features and imposes $O(p)$ (orthogonal group) equivariance w.r.t. the positional features simultaneously where $p$ is the dimension of used positional features. Extensive link prediction experiments over 8 real-world networks demonstrate the advantages of \proj in generalization and scalability.\footnote{Code available at \url{https://github.com/Graph-COM/PEG}}

\end{abstract}
%\vspace{-4.5mm}
\section{Introduction}
%\vspace{-2.5mm}
Graph neural networks (GNN), inheriting from the power of neural networks~\citep{hornik1989multilayer}, have recently become the de facto standard for machine learning with graph-structured data~\citep{scarselli2008graph}. While GNNs can easily outperform traditional algorithms for single-node tasks (such as node classification) and whole graph tasks (such as graph classification), GNNs predicting over a set of nodes often achieve subpar performance. For example, for link prediction, GNN models, such as GCN~\citep{kipf2016semi}, GAE~\citep{kipf2016variational} may perform even worse than some simple heuristics such as common neighbors and Adamic Adar~\citep{liben2007link} (see the performance comparison over the networks Collab and PPA in Open Graph Benchmark (OGB)~\citep{hu2020open}). Similar issues widely appear in node-set-based tasks such as network motif prediction~\citep{liu2022neural,besta2021motif}, motif counting~\citep{zhengdao2020can}, relation prediction~\citep{wang2021relational,teru2020inductive} and temporal interaction prediction~\citep{wang2021inductive}, which posts a big concern for applying GNNs to these relevant real-world applications.   

%graph auto-encoder (GAE)~\citep{kipf2016variational}

\begin{figure}[t]
\centering
    \includegraphics[trim={0.5cm 15.7cm 3.3cm 2.6cm},clip,width=1.0\textwidth]{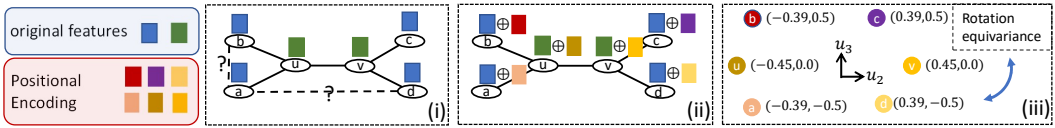}
    %\vspace{-0.7cm}
    \caption{\small{Illustration of Positional Encoding (PE): (i) GNNs cannot distinguish nodes a,b,c,d because the graph has automorphism where $a$ is mapped to $b,c,d$. GNNs fail to predict whether (a,b) or (a,d) is more likely to have a link; (ii) PE associates each node with extra positional features that may distinguish nodes; (iii) An example of PE uses the eigenvectors that correspond to the 2nd and 3rd smallest eigenvalues of the graph Laplacian as positional features (denoted as the 2-dim vectors besides each node). The proposed GNN layer \proj keeps $O(2)$ equivariance when processing these features.}}
     %\vspace{-0.5cm}
    \label{fig:diagram}
\end{figure}

The above failure is essentially induced by the loss of node identities during the intrinsic computation of GNNs. Nodes that get matched under graph automorphism will be associated with the same representation by GNNs and thus are indistinguishable (see Fig.~\ref{fig:diagram}(i)). A naive way to solve this problem is to pair GNNs with one-hot encoding as the extra node feature. However, it violates the fundamental inductive bias, i.e., permutation equivariance which GNNs are designed for, and thus may lead to poor generalization capability: The obtained GNNs  are not transferable (inductive) across different node sets and different graphs or stable to network perturbation. 

Many works have been recently proposed to address such an issue. The key idea is to use augmented node features, where either random features (RF) or deterministic distance encoding (DE) can be adopted.
Interested readers may refer to the book chapter~\citep{GNNBook-ch5-li} for detailed discussion. Here we give a brief review. RF by nature distinguishes nodes and guarantees permutation equivariance if the distribution to generate RF keep invariant across the nodes. Although GNNs paired with RF have been proved to be more expressive~\citep{murphy2019relational,sato2021random}, the training procedure is often hard to converge and the prediction is noisy and inaccurate due to the injected randomness~\citep{abboud2020surprising}. %Their empirical performances for node-set-based tasks are generally not ideal. 
On the other hand, DE defines extra features by using the distance from a node to the node set where the prediction is to be made~\citep{li2020distance}. %For each node set to be predicted over, the relative distance from the node set to a node in the graph is used as the node extra feature~\citep{}. 
This technique is theoretically sound and empirically performs well~\citep{zhang2018link,li2020distance,zhang2020revisiting}. But it introduces huge memory and time consumption. This is because DE is specific to each node set sample and no intermediate computational results, e.g., node representations in the canonical GNN pipeline can be shared across different samples. 

%Some works combine the above two types of techniques via injecting random node positional encoding. 

To alleviate the computational cost of DE, absolute positions of nodes in the graph may be used as the extra features. We call this technique as  positional encoding (PE). PE may approximate DE by measuring the distance between positional features and can be shared across different node set samples. % and thus compensate for the above DF technique. 
%PE can be shared across different samples and thus avoid the huge memory and time consumption. 
%GNN using PE is not a novel concept, but
However, the fundamental challenge is how to guarantee the GNNs trained with PE keep permutation equivariant and stable. Using the idea of RF, previous works randomize PE to guarantee permutation equivariance. %\muhan{I think PGNN is not inspired by RF. They are simultaneous works. And PGNN did not intend to solve the permutation equivariant problem. Only if the random anchors are repeatedly sampled will PGNN be permutation equivariant.} \pan{PGNN indeed uses repeatedly sampling; The idea of RF, random data augmentation, is earlier than GNN+RF. I refer to the big idea of RF.} 
Specifically, \citet{you2019position} designs PE as the distances from a node to some randomly selected anchor nodes. However, the approach suffers from slow convergence and achieves merely subpar performance. \citet{srinivasan2020equivalence} states that PE using the eigenvectors of the randomly permuted graph Laplacian matrix keeps permutation equivariant. \citet{dwivedi2020generalization,kreuzer2021rethinking} argue that such eigenvectors are unique up to their signs and thus propose PE that randomly perturbs the signs of those eigenvectors. Unfortunately, these methods may have risks. They cannot provide permutation equivariant GNNs when the matrix has multiple eigenvalues, which thus are dangerous when applying to many practical networks. For example, large social networks, when not connected, have multiple 0 eigenvalues; small molecule networks often have non-trivial automorphism that may give multiple eigenvalues.
%\muhan{Most graphs will have distinct eigenvalues. I think only regular graphs have identical eigenvalues?}  \pan{More than that, disconnected graphs will have multiple 0 eigenvalues. More severely, stability is more critical.}\muhan{I see. Then you mean have eigenvalues with multiplicity larger than 1. ``multiple eigenvalues'' was initially confusing to me.} . 
Even if the eigenvalues are distinct, these methods are unstable. We prove that the sensitivity of node representations  to the graph perturbation depends on the inverse of the smallest gap between two consecutive eigenvalues, which could be actually large when two eigenvalues are close (Lemma~\ref{thm:unstable}).%\muhan{The discussion on eigenvectors is confusing to me. Is it too difficult to put in introduction?} \pan{I am to revise here somehow later.}

In this work, we propose a principled way of using PE to build more powerful GNNs. \emph{The key idea is to use separate channels to update the original node features and positional features. The GNN architecture not only keeps permutation equivariant w.r.t. node features but also keeps rotation and reflection equivariant w.r.t. positional features.} This idea applies to a broad range of PE techniques that can be formulated as matrix factorization~\citep{qiu2018network} such as Laplacian Eigenmap (LE)~\citep{belkin2003laplacian} and Deepwalk~\citep{perozzi2014deepwalk}. We design a GNN layer \proj that satisfies such requirements. \proj is provably stable: In particular, we prove that the sensitivity of node representations learnt by \proj only depends on the gap between the $p$th and $(p+1)$th eigenvalues of the graph Laplacian if $p$-dim LE is adopted as PE, instead of the smallest gap between any two consecutive eigenvalues that previous works have achieved.

\proj gets evaluated in the most important node-set-based task, link prediction, over 8 real-world networks. \proj achieves comparable performance with strong baselines based on DE while having much lower training and inference complexity. \proj achieves significantly better performance than other baselines without using DE. Such performance gaps get enlarged when we conduct domain-shift link prediction, where the networks used for training and testing are from different domains, which effectively demonstrates the strong generalization and transferability of \proj.

%In transductive link prediction is to evaluate the power of PE-GNN, where PE-GNN outperforms .... . Inductive link prediction is to evaluate stability and generalization of PE-GNN, where xxx.   
%\vspace{-3mm}
\subsection{Other related works}
%\vspace{-2mm}
%Canonical GNNs generally follow a node-feature-refinement procedure where a node updates its self features by combining with the features from its neighbors~\citep{hamilton2017inductive,gilmer2017neural,morris2019weisfeiler}. 
As long as GNNs can be explained by a node-feature-refinement procedure~\citep{hamilton2017inductive,gilmer2017neural,morris2019weisfeiler,velivckovic2018graph,klicpera2019predict,chien2021adaptive}, they suffer from the aforementioned node ambiguity issue. Some GNNs cannot be explained as node-feature refinement as they directly track node-set representations~\citep{maron2018invariant,morris2019weisfeiler,chen2019equivalence,maron2019provably}. However,
their complexity is high, as they need to compute the representation of each node set of certain size. Moreover, they were only studied for graph-level tasks. EGNN~\citep{satorras2021n} seems to be a relevant work as it studies when the nodes have physical coordinates given in prior. However, no analysis of PE has been provided. %Hence, how PE affects the permutation equivariance and stability of the GNNs has not been investigated.

A few works studied the stability of GNNs by using tools like graph scattering transform (for graph-level representation)~\citep{gama2019diffusion,gama2019stability} and graph signal filtering (for node-level representations)~\citep{levie2021transferability,ruiz2020graphon,ruiz2021graph,gama2020stability,nilsson2020experimental}. They all focus on justifying the stability of the canonical GNN pipeline, graph convolution layers in particular. None of them consider positional features let alone the stability of GNNs using PE.  %in graph representation. 

\section{Notations and Preliminaries}
%\vspace{-2mm}
%This section introduces relevant materials about graph neural network and permutation equivariance, which will be utilized in the description of our method.

%\pan{Notations: You need to have enough notation setup to introduce GNN and permutation equivariance. Also, discuss a little bit why permutation equivariance is important.}
%\pan{I do not think you need to introduce translation \& rotation equivariance. We just focus on orthogonal group.}

%\pan{Define degree $D$, define expectation, define eigenvalue decomposition, and }

In this section, we prepare the notations and preliminaries that are useful later. First, we define graphs.

\begin{definition}[Graph] Unless specified, we always consider undirected graphs of $N$ nodes and let $[N]=\{1,2,...,N\}$. One such graph can be denoted as $\gG=(A,X)$, where $A$ is the adjacency matrix. $X\in\mathbb{R}^{N\times F}$ denotes the node features, where the $v$th row, $X_v$, is the feature vector of node $v$. A graph may have self loops, i.e., $A$ has nonzero diagonals. Denote $D$ as the diagonal degree matrix where $D_{vv}=\sum_{u\in[N]}A_{vu}$ for $v\in[N]$. Let $d_{\max}=\max_{v\in[N]}D_{vv}$. %Denote a degree normalized adjacency matrix with a self-loop as $\hat{A} = (A + I)$
Denote the normalized adjacency matrix as $\hat{A}=D^{-\frac{1}{2}}AD^{-\frac{1}{2}}$ and the normalized Laplacian matrix as $L=I-\hat{A}$.
\end{definition} 

\begin{definition}[Permutation] An $N$-dim permutation matrix $P$ is a matrix in $\{0,1\}^{N\times N}$ where each row and each column has only one single 1. All such matrices are collected in $\Pi(N)$, simplified as $\Pi$.
\end{definition} %\vspace{-1mm}

We denote the vector $\ell_2$-norm as $\|\cdot\|$, the Frobenius norm as $\|\cdot\|_\text{F}$ and the operator norm as $\|\cdot\|_\text{op}$. %to measure the matrix distance. 

\begin{definition}[Graph-matching] Given two graphs $\gG^{(i)}=(A^{(i)},X^{(i)})$ and their normalized Laplacian matrices $L^{(i)}$ for $i\in\{1,2\}$, their matching can be denoted by a permutation matrix $P\in \Pi$ that best aligns the graph structures and the node features.  %\vspace{-0.5mm}
\begin{align*}
    P^{*}(\gG^{(1)}, \gG^{(2)}) \triangleq \argmin_{P\in \Pi}  \|L^{(1)} - PL^{(2)}P^T\|_{\text{F}} + \| X^{(1)} - PX^{(2)}\|_{\text{F}}
\end{align*}
Using $L$ instead of $A$ to represent graph structures is for notational simplicity. Actually for an unweighted graph, there is a bijective mapping between $L$ and $A$. One can rewrite the first term with $A$. Later, we use $P^*$ by not specifying $\gG_1$ and $\gG_2$ if there is no confusion. The distance between the two graphs can be defined as $d(\gG_1, \gG_2) = \|L^{(1)} - P^*L^{(2)}P^{*T}\|_{\text{F}} + \| X^{(1)} - P^*X^{(2)}\|_{\text{F}}$.
\end{definition}

Next, we review eigenvalue decomposition and summarize arguments on its uniqueness in Lemma~\ref{thm:evd}. 

\begin{definition}[Eigenvalue Decomposition (EVD)] \label{def:evd}
For a positive semidefinite (PSD) matrix $B\in \mathbb{R}^{N\times N}$, it has eigenvalue decomposition $B=U\Lambda U^T$ where $\Lambda$ is a real diagonal matrix with the eigenvalues of $B$, $0\leq \lambda_1 \leq \lambda_2\leq ... \leq \lambda_N$ as its diagonal components. $U = [u_1, u_2,...,u_N]$ is an orthogonal matrix where $u_i\in\mathbb{R}^N$ is the $i$th eigenvector, i.e. $Bu_i =\lambda_i u_i$.
\end{definition}

\begin{definition}[Orthogonal Group in the Euclidean space]
$\text{O}(k) =\{Q\in \mathbb{R}^{k\times k}| QQ^T = Q^TQ= I\}$ includes all $k$-by-$k$ orthogonal matrices. A subgroup of $\text{O}(k)$ includes all diagonal matrices with $\pm1$  as the diagonal components, i.e., reflection $\text{SN}(k) = \{Q\in \mathbb{R}^{k\times k}| Q_{ii}\in\{-1,1\}, Q_{ij} = 0, \, \text{for}\, i\neq j\}$.
\end{definition}
%$I$ is the identity matrix.
%\textbf{Definition 2.4} (Orthogonal group) A $n \times n$ matrix is an orthogonal matrix if $AA^T = I$, where $A^T$ is the transpose of $A$ and $I$ is the identity matrix. For every $n>0$, since the $n \times n$ orthogonal matrices are closed under multiplication and taking inverses, they form the orthogonal group $SO(n)$. 

%Note that EVD is not unique due to the following theorem.

\begin{lemma} \label{thm:evd}
 EVD is not unique. If all the eigenvalues are distinct, i.e., $\lambda_i\neq \lambda_j$, $U$ is unique up to the signs of its columns, i.e., replacing $u_i$ by $-u_i$ also gives EVD. If there are multiple eigenvalues, say $(\lambda_{i-1}<)\lambda_i = \lambda_{i+1} =...=\lambda_{i+k-1} (< \lambda_{i+k})$, then $[u_i, u_{i+1}, ...,u_{i+k-1}]$ lie in an orbit induced by the orthogonal group $\text{O}(k)$, i.e., replacing $[u_i, u_{i+1}, ...,u_{i+k-1}]$ by $[u_i, u_{i+1}, ...,u_{i+k-1}]Q$ for any $Q\in \text{O}(k)$ while keeping eigenvalues and other eigenvectors unchanged also gives EVD.
\end{lemma}

Next, we define Positional Encoding (PE), which associates each node with a vector in a metric space where the distance between two vectors can represent the distance between the nodes in the graph. %Traditional network embeddings all provide effective positional encoding techniques~\citep{belkin2003laplacian,tang2015line,perozzi2014deepwalk,grover2016node2vec}. 
\begin{definition}[Positional Encoding]
Given a graph $\gG = (A, X)$, PE works on $A$ and gives $Z=\text{PE}(A)\in\mathbb{R}^{N\times p}$ where each row $Z_v$ gives the positional feature of node $v$.
\end{definition} %\vspace{-0.5mm}
The absolute values given by PE may not be useful while the distances between the positional features are more relevant. So, we define PE-matching that allows rotation and reflection (orthogonal operations) to best match positional features, which further defines the distance between two collections of positional features. %\vspace{-1mm} %Using rotation to match positional features implies that the inner product kernel   %Later, we use $\mathbf{1}\in\mathbb{R}^{N}}$ to denote the all 1 vector and by default assume all vectors are vertical. 
%For most downstream tasks, the absolute values of positional features may not useful while the distances between the positional features are more relevant. To measure the distance between two groups of positional features, we need to define PE-matching that allows rotate and translate positional features.    
\begin{definition}[PE-matching] \label{def:pe-matching}
Consider two groups of positional features $Z^{(1)},\, Z^{(2)}\in\mathbb{R}^{N\times p}$. Their matching is given by $Q^*(Z^{(1)}, Z^{(2)})\triangleq \argmin_{Q\in \text{O}(p)}\|Z^{(1)} - Z^{(2)}Q \|_{\text{F}}$. Later, $Q^*$ is used if it causes no confusion. Define the distance between them as $\eta(Z^{(1)}, Z^{(2)}) = \|Z^{(1)} - Z^{(2)}Q^*\|_{\text{F}}$.  
\end{definition}

\section{Equivariant and Stable Positional Encoding for GNN}
%\vspace{-2mm}
In this section, we will study the equivariance and stability of a GNN layer using PE . %First, let us formally define the goal that is to be achieved.

%\vspace{-3mm}
\subsection{Our Goal: Building Permutation Equivariant and Stable GNN layers}
%\vspace{-2mm}

A GNN layer is expected to be permutation equivariant and stable. These two properties, if satisfied, guarantee that the GNN layer is transferable and has better generalization performance. Permutation equivariance implies that model predictions should be irrelevant to how one indexes the nodes, which captures the fundamental inductive bias of many graph learning problems: A permutation equivariant layer can make the same prediction over a new testing graph as that over a training graph if the two graphs match each other perfectly, i.e. the distance between them is 0. Stability is actually an even stronger condition than permutation equivariance because it characterizes how much gap between the predictions of the two graphs is expected when they do not perfectly match each other. %can be well controlled.  %If a GNN layer is permutation equivariant, it is expected to make the same prediction when it is applied to a new graph for test, which has never been seen during the training. . 

Specifically, given a graph $A$ with %node representations $H_{in}\in\mathbb{R}^{N\times F}$, 
node features $X$, we consider a GNN layer $\tilde{g}$ updating the node features, denoted as $\hat{X} = \tilde{g}(A,X)$. %\in \mathbb{R}^{N\times F'}$.  %where $W$ corresponds to learnable parameters. %GNN layers are expected to satisfies permutation equivariance, which implies how to index the nodes is irrelevant.
We define permutation equivariance and stability as follows. 
\begin{definition}[Permutation Equivariance]
A GNN layer $\tilde{g}$ is permutation equivariant, if for any $P\in \Pi$ and any graph $\gG=(A, X)$, $P\tilde{g}(A,X) = \tilde{g}(PAP^T,PX)$.
\end{definition}%\vspace{-1mm}

\begin{definition}[Stability]
A GNN layer $\tilde{g}$ is claimed to be stable, if there exists a constant $C>0$, for any two graphs $\gG^{(1)}=(A^{(1)}, X^{(1)})$ and $\gG^{(2)} = (A^{(2)}, X^{(2)})$, letting $P^{*} = P^{*}(\gG^{(1)}, \gG^{(2)})$ denote their matching, $\tilde{g}$ satisfies  
%\begin{align} \label{eq:stability}
    $\| \tilde{g}(A^{(1)}, X^{(1)}) - P^*\tilde{g}(A^{(2)}, X^{(2)})\|_{\text{F}} \leq Cd(\gG^{(1)}, \gG^{(2)})$. 
%\end{align}
By setting $A^{(1)} = PA^{(2)}P^T$ and $X^{(1)} = PX^{(2)}$ for some $P\in \Pi$, the RHS becomes zero, so a stable $\tilde{g}$ makes the LHS zero too. So, stability is a stronger condition than permutation equivariance.
\end{definition}%\vspace{-1mm}

Our goal is to guarantee that the GNN layer that utilizes PE  is permutation equivariant and stable. %Positional encoding (PE) is the class of techniques to associate each node in the graph with a dense vector where the distance between two vectors can approximate the distance between their corresponding nodes in the graph. 
%Given a graph $\gG = (A, X)$ with $N$ nodes, PE works on $A$ and gives $Z=\text{PE}(A)\in\mathbb{R}^{N\times p}$ where each row $Z_v$ gives the positional feature of node $v$. %The GNN layer that uses $Z$ is denoted as $g(A,X,Z) = g(A,X,\text{PE}(A))$.
To distinguish from the GNN layer $\tilde{g}$ that does not use PE, we use $g$ to denote a GNN layer that uses PE, which takes the positional features $Z$ as one input and may update both node features $X$ and  $Z$, i.e., $(\hat{X},\hat{Z})= g(A,X,Z)$. 
%We define a GNN layer $g$ that uses PE may update both node features $X$ and positional features, i.e., $(X,Z)= g(A,X,Z)$
Now, we may define PE-equivariance and PE-stability for the GNN layer $g$.

\begin{definition}[PE-stability \& PE-equivariance]
Consider a GNN layer $g$ that uses PE. When it works on any two graphs $\gG^{(i)} = (A^{(i)}, X^{(i)})$, $i\in \{1,\,2\}$ and gives $(\hat{X}^{(i)},\hat{Z}^{(i)}) = g(A^{(i)}, X^{(i)}, Z^{(i)})$, let $P^{*}$ be the matching between the two graphs. $g$ is PE-stable, if for some constant $C>0$ we have
\begin{align} \label{eq:pe-stability}
        \| \hat{X}^{(1)} - P^*\hat{X}^{(2)}\|_{\text{F}} + \eta(\hat{Z}^{(1)}, P^*\hat{Z}^{(2)}) \leq Cd(\gG^{(1)}, \gG^{(2)}).
\end{align}
Recall $\eta(\cdot,\cdot)$ measures the distance between two sets of positional features as defined in Def. \ref{def:pe-matching}.
Similar as above, a weaker condition of PE-stability is PE-equivariance: If $A^{(1)} = PA^{(2)}P^T$ and $X^{(1)} = PX^{(2)}$ for some $P\in \Pi$, we expect a perfect match between the updated node features and positional features, $\hat{X}^{(1)} = P\hat{X}^{(2)}$ and $\eta(\hat{Z}^{(1)}, P\hat{Z}^{(2)})=0$.
\end{definition}
%\vspace{-1mm}
Note that previous works also consider $g$ that updates only node features, i.e., $\hat{X}= g(A,X,Z)$. In this case, PE-stability can be measured by removing the second term on $\hat{Z}$ from Eq.\ref{eq:pe-stability}.

% \begin{definition}[PE-Stability \& PE-Equivariance]
% A GNN layer $g$ that uses PE is PE-stable, if it works on any two graphs $\gG^{(1)}=(A^{(1)}, X^{(1)})$ and $\gG^{(2)} = (A^{(2)}, X^{(2)})$, letting $P^{*} = P^{*}(\gG^{(1)}, \gG^{(2)})$ denote their matching, then for some constant $C$, 
% \begin{align} \label{eq:pe-stability}
%     \| g(A^{(1)}, X^{(1)}, \text{PE}(A^{(1)})) - P^*g(A^{(2)}, X^{(2)}, \text{PE}(A^{(2)}))\| \leq Cd(\gG^{(1)}, \gG^{(2)})
% \end{align}
% Similarly as above, a weaker condition of PE-Stability is PE-Equivariance, i.e., $Pg(A,X, \text{PE}(A)) = g(PAP^T,PX,\text{PE}(PAP^T))$ for any $P\in \Pi$.
% \end{definition}
%\vspace{-2mm}
\subsection{PE-stable GNN layers based on Laplacian Eigenmap as Positional Encoding}
%\vspace{-1mm}
\begin{figure}[t]
\centering
    \includegraphics[trim={2.9cm 0.3cm 4.0cm 0.9cm},clip,width=0.46\textwidth]{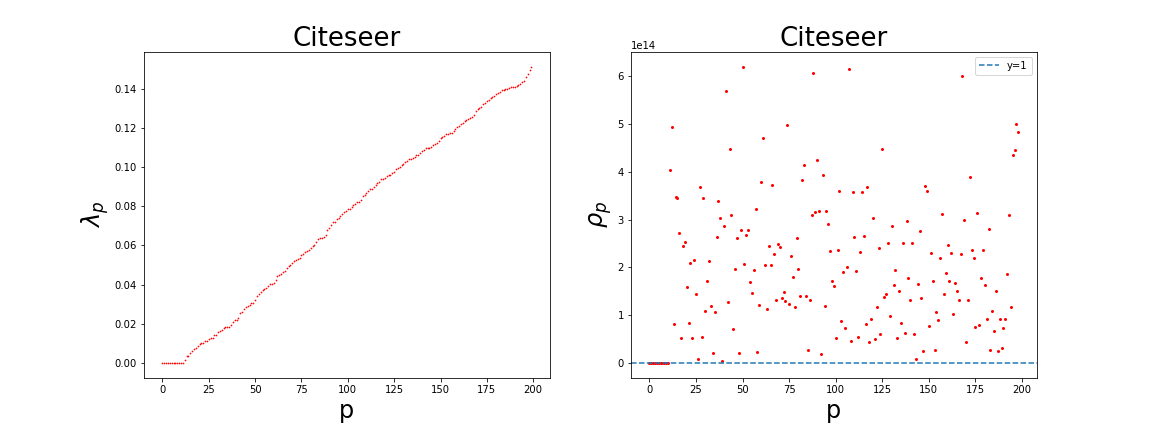}
    \includegraphics[trim={2.9cm 0.3cm 4.0cm 0.9cm},clip,width=0.46\textwidth]{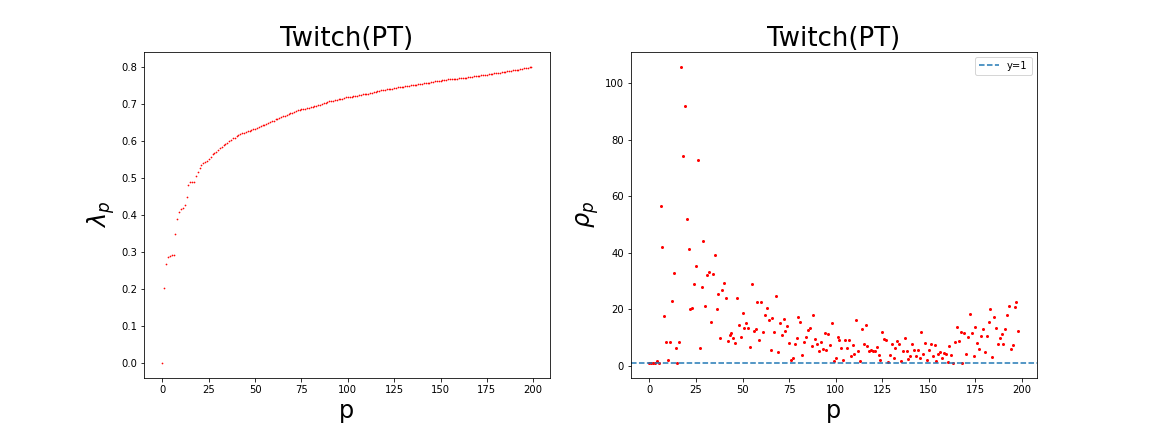}
    %\vspace{-0.3cm}
    \caption{\small{Eigenvalues $\lambda_p$ and the stability ratio $\rho_p=\frac{|\lambda_p - \lambda_{p+1}|}{\min_{1\leq k\leq p}|\lambda_k - \lambda_{k+1}|}$. \proj with sensitivity $|\lambda_p - \lambda_{p+1}|^{-1}$  is far more stable than previous methods with sensitivity $\max_{1\leq k\leq p}|\lambda_k - \lambda_{k+1}|^{-1}$. Over Citeseer, $\rho_p$ is extremely large because there are multiple eigenvalues ($\rho_p$ is still finite due to the numerical approximation of the eigenvalues). Over Twitch(PT), even if there are no multiple eigenvalues, $\rho_p$ is mostly larger than 10. }}
     %\vspace{-0.3cm}
    \label{fig:eigengaps}
\end{figure}

To study the requirement of the GNN layer that achieves PE-stability, let us start from a particular PE technique, i.e., Laplacian eigenmap (LE)~\citep{belkin2003laplacian} to show the key insight behind. Later, we generalize the concept to some other PE. Let $Z_{LE}$ denote LE, which includes the eigenvectors that correspond to the $p$ smallest eigenvalues of the normalized Laplacian matrix $L$. %(or the Laplacian matrix $\hat{L}\triangleq D - A$).

\textbf{Previous works failed to design PE-equivariant or PE-stable GNN layers.} \citet{srinivasan2020equivalence} claims that if a GNN layer $\tilde{g}(A,X)$ that does not rely on PE is permutation equivariant, PE-equivariance may be kept by adding $Z_{LE}$ to node features, i.e., $g(A,X,Z_{LE})=\tilde{g}(A, X + \text{MLP}(Z_{LE}))$, where MLP$(\cdot)$ is a multilayer perceptron that  adjusts the dimension properly. This statement is problematic when $Z_{LE}$ is not unique given the graph structure $A$. Specifically, though sharing graph structure $A^{(1)}=A^{(2)}$, if different implementations lead to different LEs $Z_{LE}^{(1)}\neq Z_{LE}^{(2)}$, then $\tilde{g}(A^{(1)}, X^{(1)} + \text{MLP}(Z_{LE}^{(1)})) \neq \tilde{g}(A^{(2)}, X^{(2)} + \text{MLP}(Z_{LE}^{(2)}))$, which violates PE-equivariance. \citet{srinivasan2020equivalence} suggests using graph permutation augmentation to address the issue, which makes assumptions on an invariant distribution of $Z_{LE}$ that may  not be guaranteed empirically. 

\citet{dwivedi2020generalization,kreuzer2021rethinking} claim the uniqueness of $Z_{LE}$ up to the signs and suggest building a GNN layer that uses random $Z_{LE}$ as $g(A, X, Z_{LE}) = \tilde{g}(A, X+ \text{MLP}(Z_{LE}S))$ where $S$ is uniformly at random sampled from $\text{SN}(p)$. They expect PE-equivariance in the sense of expectation. %. i.e., $P\mathbb{E}[\tilde{g}(A,X+\text{MLP}(Z_{LE}S))] = \mathbb{E}[\tilde{g}(PAP^T,PX+\text{PE}(PAP^T)S)]$ for any $P\in \Pi$. 
However, this statement is generally incorrect because it depends on the condition that \emph{all eigenvalues have to be distinct} as stated in Lemma~\ref{thm:evd}. %Many practical networks have a Laplacian $L$ with multiple eigenvalues. For example, large social networks are generally not connected, so there are multiple 0 eigenvalues; small molecule networks often have non-trivial automorphism that gives multiple eigenvalues. 
Actually, for multiple eigenvalues, there are infinitely many eigenvectors that lie in the orbit induced by the orthogonal group. Although many real graphs have multiple eigenvalues such as disconnected graphs or graphs with some non-trivial automophism, 
 %(see Theorem~\ref{thm:evd})
one may argue that the methods work when the eigenvalues are all distinct. However, the above failure may further yield PE-instability even when the eigenvalues are distinct but have small gaps due to the following lemma.

\begin{lemma} \label{thm:unstable}
For any PSD matrix $B\in \mathbb{R}^{N\times N}$ without multiple eigenvalues, set positional encoding $\text{PE}(B)$ as the eigenvectors given by the smallest $p$ eigenvalues sorted as $0=\lambda_1 < \lambda_2 <...<\lambda_p (<\lambda_{p+1})$ of $B$. For any sufficiently small $\epsilon>0$, there exists a perturbation $\triangle B$, $\|\triangle B\|_{\text{F}}\leq \epsilon$ such that
\begin{align}
    \min_{S\in \text{SN}(p)}\|\text{PE}(B) - \text{PE}(B+\triangle B)S\|_F \geq 0.99\max_{1\leq i \leq p} |\lambda_{i+1} -\lambda_{i}|^{-1}\|\triangle B\|_{\text{F}} + o(\epsilon).
\end{align}
%Define $Z = \text{PE}(B)$. %= [v_1, v_2,...,v_Q]$. Then, there exists a small perturbation $\triangle B$, $B'=B+\triangle$ yields new positional features $Z = \text{PE}(B') = [v'_1, v'_2,...,v_Q]$  
\end{lemma}
%\vspace{-1mm}
Lemma \ref{thm:unstable} implies that small perturbation of graph structures may yield a big change of eigenvectors if there is a small eigengap. Consider two graphs $\gG^{(1)}=(A^{(1)}, X^{(1)})$ and $\gG^{(2)} = (A^{(2)}, X^{(2)})$, where $X^{(1)} = X^{(2)}$ and $A^{(2)}$ is $A^{(1)}$ with a small perturbation $\triangle A^{(1)}$. The perturbation is small enough so that the matching $P^*(\gG^{(1)}, \gG^{(2)}) = I$. However, the change of PE even after removing the effect of changing signs $\min_{S\in \text{SN}(p)}\|\text{PE}(A^{(1)}) - \text{PE}(A^{(2)})S\|$ could be dominated by the largest inverse eigengap among the first $p+1$ eigenvalues $\max_{1\leq i \leq p} |\lambda_{i+1} -\lambda_{i}|^{-1}$. In practice, it is hard to guarantee all these $p$ eigenpairs have large gaps, especially when a large $p$ is used to locate each node more accurately. Plugging this PE into a GNN layer gives the updated node features $\hat{X}^{(i)}$ a large change  $\|\hat{X}^{(1)} - \hat{X}^{(2)}\|_\text{F}$ and thus violates PE-stability.

Note that although the way to use PE in previous works \citep{dwivedi2020generalization,kreuzer2021rethinking} and other concurrent works~\citep{lim2022sign,dwivedi2022graph} has the risk of being unstable, their models seem to work fine in their experiments, which is due to the experiment setting. They focus on the tasks where the datasets contain a large number ($\geq 1000$) of small graphs (at $\leq 100-200$ nodes per graph) for training, while our experiments in Sec.~\ref{sec:exp} focus on the tasks where the datasets contain a small number ($\leq 10$) of large graphs ($\geq 1000$ nodes per graph). %In this case, the risk of the model instability gets diminished.
\emph{We conjecture that the instability issue of the models become more severe when the graphs are of larger sizes and the training set is smaller.} We leave an extensive study of this point in the future. %Our experiments in Sec.~\ref{sec:exp} test over the tasks with a small number of large graphs.
 
Note that even in the case with a large number of small graphs for training, \citet{dwivedi2022graph} also observes the instability of using sign-perturbed LE and thus proposes to set PE as a variant of deterministic distance encoding~\citep{li2020distance}, the random walk landing probabilities from one node to itself. As distance encoding does not need to involve matrix factorization, it does not suffer from the instability issue discussed in this paper. Graphomer~\citep{ying2021transformers} also adopts a type of distance encoding, shortest path distance between two nodes as edge features, instead of positional encoding so it also does not have the instability issue discussed in the paper. %In this paper, we focus on the PE obtained from matrix factorization as used in \citep{srinivasan2020equivalence,dwivedi2020generalization,kreuzer2021rethinking,lim2022sign}. We argue that the instability of the models becomes more obvious when the graphs are of larger sizes and the training set is smaller. Our experiments in Sec.~\ref{sec:exp} adopt the tasks over a small number of large graphs.

%\begin{align}
%    \min_{S\in SN(p)}\|g(A^{(1)}, X^{(1)}+ \text{MLP}(PE(A^{(1)}))) - P^*g(A^{(2)}, X^{(2)}+ \text{MLP}(PE(A^{(2)})S))\| \geq \max_{1\leq i \leq p} |\lambda_{i+1} -\lambda_{i}|^{-1} d(\gG^{(1)}, \gG^{(2)})
%\end{align}

\textbf{PE-stable GNN layers.} 
%is expected to satisfy $g(A, X, \text{PE}(A)) = g(A, X, \text{PE}(A)Q)$ for all $Q\in S\text{O}(p)$. 
Although a particular eigenvector may be not stable, the eigenspace, i.e., the space spanned by the columns of $\text{PE}(A)$ could be much more stable. This motivates our following design of the \textit{PE-stable GNN layers}. Formally, we use the following lemma that can characterize the distance between the eigenspaces spanned by LEs of two graph Laplacians. The error is controlled by the inverse eigengap between the $p$th and $(p+1)$th eigenvalues $\min_{i=1,2}\{|\lambda_p^{(i)} - \lambda_{p+1}^{(i)}|^{-1}\}$, which by properly setting $p$ is typically much smaller than $\max_{1\leq k\leq p} |\lambda_k - \lambda_{k+1}|^{-1}$ in Lemma \ref{thm:unstable}. We compute the ratio between these two values over some real-world graphs as shown in Fig.~\ref{fig:eigengaps}.

%implies that for two graph Laplacians, the distance between the eigenspaces constructed by their LEs can be controlled by the inverse eigengap between the $p$th and $p+1$th eigenvalues $\min_{i=1,2}\{|\lambda_p^{(i)} - \lambda_{p+1}^{(i)}|^{-1}\}$, which requires much less than $\max_{p} |\lambda_p - \lambda_{p+1}|^{-1}$ in Thm.~\ref{thm:unstable}.   
\begin{lemma} \label{thm:stable}
For two PSD matrices $B^{(1)},\,B^{(2)}\in \mathbb{R}^{N\times N}$, set $\text{PE}(B)$ as the eigenvectors given by the $p$ smallest eigenvalues of $B$. Suppose $B^{(i)}$ has eigenvalues $0=\lambda_1^{(i)}\leq \lambda_2^{(i)}\leq...\leq\lambda_p^{(i)}\leq \lambda_{p+1}^{(i)}$ and $\delta = \min_{i=1,2}\{|\lambda_p^{(i)} - \lambda_{p+1}^{(i)}|^{-1}\}<\infty$. Then, for any permutation matrix $P\in \Pi$,
%\small{
\begin{align}
    \eta(\text{PE}(B^{(1)}), P\text{PE}(B^{(2)})) \leq 2^{\frac{3}{2}}\delta \min\{\sqrt{p}\|B^{(1)} - PB^{(2)}P^{T}\|_\text{op}, \|B^{(1)} - PB^{(2)}P^{T}\|_\text{F}\} 
\end{align}
%In particular, if $p=N$, $\text{PE}(B)\text{PE}(B)^T = \text{PE}(B+\triangle B)\text{PE}(B+\triangle B)^T$
%Define $Z = \text{PE}(B)$. %= [v_1, v_2,...,v_Q]$. Then, there exists a small perturbation $\triangle B$, $B'=B+\triangle$ yields new positional features $Z = \text{PE}(B') = [v'_1, v'_2,...,v_Q]$  
%
\end{lemma}
%\vspace{-1mm}
Inspired by the stability of the eigenspace, the idea to achieve PE-stability is to make the GNN layer invariant to the selection of bases of the eigenspace for the positional features. So, our proposed PE-stable GNN layer $g$ that uses PE should satisfy two necessary conditions: 1) Permutation equivariance w.r.t. all features; 2) $\text{O}(p)$ (rotation and reflection) equivariance w.r.t. positional features, i.e.,
\begin{align}
& \textbf{PE-stable layer cond. 1:}    &(P\hat{X},P\hat{Z}) &= g(PAP^T,PX,PZ), \;\forall P\in \Pi(N), \label{eq:pi-eq}\\
& \textbf{PE-stable layer cond. 2:}  &(\hat{X},\hat{Z}Q) &= g(A,X,ZQ), \;\forall Q\in \text{O}(p). \label{eq:oth-eq}
\end{align}
%$g(A, X, \text{PE}(A)) = g(A, X, \text{PE}(A)Q)$, $\forall Q\in SO(p)$. 

The $\text{O}(p)$ equivariance reflects the eigenspace instead of a particular selection of eigenvectors and thus achieves much better stability. Interestingly, these requirements can be satisfied by EGNN recently proposed~\citep{satorras2021n} as one can view the physical coordinates of objects considered by EGNN as the positional features. EGNN gets briefly reviewed in Appendix~\ref{apd:EGNN}. %Actually, EGNN asks more by also requiring translation equivariance w.r.t. positional features. However, translation equivariance is unnecessary for a GNN to achieve PE-stability \& PE-equivariance because the positional features given by PE are generally not in the Euclidean space as considered by EGNN. E.g., LE lies on a spherical space.
Thm.~\ref{thm:equivariant} proves PE-equivariance under the conditions Eqs.~\ref{eq:pi-eq} and \ref{eq:oth-eq}. %of a GNN layer when Eqs.~\ref{eq:pi-eq},\ref{eq:oth-eq} are satisfied.

\begin{theorem} \label{thm:equivariant}
A GNN layer $g(A,X,Z)$ that uses $Z=Z_\text{LE}$ and satisfies Eqs.~\ref{eq:pi-eq},\ref{eq:oth-eq} is PE-equivariant if the $p$th and $(p+1)$th eigenvalues of the normalized Laplacian matrix $L$ are different, i.e., $\lambda_p\neq \lambda_{p+1}$.
\end{theorem}
%\vspace{-1mm}

%For the GNN layers considered in \cite{dwivedi2020generalization,kreuzer2021rethinking}, PE-equivariance can be guaranteed when all the $p+1$ smallest eigenvalues have to be distinct. 
Note that satisfying Eqs.~\ref{eq:pi-eq},\ref{eq:oth-eq} is insufficient to guarantee PE-stability that depends on the form of $g$.

We implement $g$ in our model PEGN with further simplification which has already achieved good empirical performance: Use a GCN layer with edge weights according to the distance between the end nodes of the edge and keep the positional features unchanged. This gives \textbf{the \proj layer}, %\normalsize
%\small{
{\small
\begin{align} \label{eq:PE-GNN-layer1}
 \textbf{PEG:}\quad g_{\text{PEG}}(A,X,Z) = \left(\psi\left[\left(\hat{A} \odot \Xi\right) X W \right], Z\right),\,\text{where $\Xi_{uv}=\phi\left(\|Z_u-Z_v\|\right)$, $\forall u,v\in[N]$}.
\end{align}}\\
\normalsize
Here $\psi$ is an element-wise activation function, $\phi$ is an MLP mapping from $\mathbb{R}\rightarrow \mathbb{R}$ and $\odot$ is the Hadamard product. Note that if $\hat{A}$ is sparse, only $\Xi_{uv}$ for an edge $uv$ needs to be computed.  
%$\hat{A}$ follows GCN by adding self-loops and degree normalization of $A$, and $\odot$ is the Hadamard product.

\textbf{Remark.} Note that the above implementation to achieve $\text{O}(p)$ equivariance is only as a proof of concept. Another way to achieve $\text{O}(p)$ equivariance is via $\Xi_{uv}=\phi\left(Z_uZ_v^T\right)$, which we find achieves similar while slightly worse empirical performance compared to Eq.~\ref{eq:PE-GNN-layer1} in the link prediction experiments in Sec.~\ref{sec:exp}.

We may also prove that the \proj layer $g_{\text{PEG}}$ satisfies the even stronger condition PE-stability. 
\begin{theorem} \label{thm:lap}
Consider two graphs $\gG^{(i)} = (A^{(i)}, X^{(i)})$, $i=1,\,2$. Denote their normalized Laplacian matrices' $p$th eigenvalue as $\lambda_p^{(i)}$ and $(p+1)$th eigenvalue as $\lambda_{p+1}^{(i)}$. Assume that $\delta = \min_{i=1,2}(\lambda_{p+1}^{(i)} - \lambda_p^{(i)})^{-1} <\infty$.   
Also, assume that $\psi$ and $\phi$ in the \proj layer in Eq.~\ref{eq:PE-GNN-layer1} are $\ell_\psi$, $\ell_\phi$-Lipschitz continuous respectively. Then, the \proj layer that uses $Z=Z_{LE}$ satisfies PE-stability with the constant $C$ in Eq.~\ref{eq:pe-stability} as $C=[(7\delta\|X^{(1)}\|_{\text{op}} + 2d_{\max}^{(2)})\ell_\psi\ell_{\phi}\|W\|_{\text{op}} + 3\delta]$.
\end{theorem}
%\vspace{-1mm}
Note that to achieve PE-stability, we need to normalize the node initial features to keep $\|X^{(1)}\|_{\text{op}}$ bounded, and control $\|W\|_{\text{op}}$ and $\ell_{\phi}$. In practice $\ell_{\psi}\leq 1$ is typically satisfied, e.g. setting $\psi$ as ReLU. Here, the most important term is $\delta$. PE-stability may only be achieved when there is an eigengap between $\lambda_p$ and $\lambda_{p+1}$ and the larger eigengap, the more stable. This observation may be also instructive to select the $p$ in practice. As previous works may encounter a smaller eigengap (Lemma~\ref{thm:unstable}), their models will be generally more unstable. 

Also, the simplified form of $g_{\text{PEG}}$ is \emph{not necessary} to achieve PE-stability. However, as it has already given consistently better empirical performance than baselines, we choose to keep using $g_{\text{PEG}}$.

%Note that the simplified form of $g_{\text{PEG}}$ may be unnecessary. But as there is no consistently better empirical performance, we choose to keep our exposition simple.% To avoid the heavy notations, we  as they are not used in our

%\vspace{-2mm}
\subsection{Generalized to Other PE techniques: DeepWalk and LINE}
%\vspace{-1mm}
It is well known that LE, as to compute the smallest eigenvalues, can be written as a low-rank matrix optimization form $Z_\text{LE}\in\argmin_{Z\in\mathbb{R}^{N\times p}} \text{tr}(L_{N}ZZ^T)$, s.t. $Z^TZ = I$. Other PE techniques, such as Deepwalk~\citep{perozzi2014deepwalk}, Node2vec~\citep{grover2016node2vec} and LINE~\citep{tang2015line} can be  unified into a similar optimization form, where the positional features $Z$ are given by matrix factorization $M^* = Z'Z^{T}$ ($M^*$ may be asymmetric so $Z'\in\mathbb{R}^{N\times p}$ may not be $Z$) and $M^*$ satisfies 
\begin{align} \label{eq:pe-unified}
 M^* = \argmin_{M\in\mathbb{R}^{N\times N}} \ell_{\text{PE}}(A, M) \triangleq\text{tr}(f_+(A)g(M)+ f_-(D) g(-M)), \quad \text{s.t.}\quad \text{rank}(M)\leq p
\end{align}
Here $f_+(\cdot):\mathbb{R}^{N\times N}\rightarrow\mathbb{R}_{\geq0}^{N\times N}$ typically revises $A$ by combining degree normalization and a power series of $A$, $f_-(\cdot):\mathbb{R}^{N\times N}\rightarrow\mathbb{R}_{\geq0}^{N\times N}$ corresponds to edge negative sampling that may be related to node degrees, and $g(\cdot)$ is a component-wise log-sigmoid function $x - \log(1+\exp(x))$. E.g., in LINE, $f_+(A) = A$, $f_-(D) = c \mathbf{1}\mathbf{1}^TD^{\frac{3}{4}}$, for some positive constant $c$, where $\mathbf{1}$ is the all-one vector. More discussion on Eq.\ref{eq:pe-unified} and the forms of $f_+$ and $f_-$ for other PE techniques are given in Appendix~\ref{apd:PE-detail}.  

According to the above optimization formulation of PE, all the PE techniques generate positional features $Z$ based on matrix factorization, thus are not unique. $Z$ always lies in the orbit induced by $\text{O}(p)$, i.e., if $Z$ solves Eq.\ref{eq:pe-unified}, $ZQ$ for $Q\in \text{O}(p)$ solves it too. A GNN layer $g$ still needs to satisfy Eqs.~\ref{eq:pi-eq},\ref{eq:oth-eq} to guarantee PE-equivariance. PE-stability actually asks even more.

%Due to the optimization in Eq.\ref{eq:pe-unified},

%The two necessary conditions of layer, as designed in Eq.\ref{eq:PE-GNN-layer1}, may handle this and achieve PE-stability and PE-equivariance given some assumptions. 

\begin{theorem}\label{thm:general-pe}
Consider a general PE$(A)$ technique that has an optimization objective as Eq.\ref{eq:pe-unified}, computes its optimal solution $M^*$  and decomposes $M^*= Z'Z^T$, s.t. $Z^TZ = I\in\mathbb{R}^{p\times p}$ to get positional features $Z$. Essentially, $Z$ consists of the right-singular vectors of $M^*$. If Eq.\ref{eq:pe-unified} has a unique solution $M^*$, $\text{rank}(M^*)=p$ and $f_+, f_-$ therein satisfy $f_+(PAP^T) = Pf_+(A)P^T$ and $f_-(PDP^T) = Pf_-(D)P^T$, then a GNN layer $g(A,X,Z)$ that satisfies Eqs.~\ref{eq:pi-eq},\ref{eq:oth-eq} is PE-equivariant. %The PE-GNN layer in Eq.~\ref{eq:pe-unified} is PE-stable with $C=...$. 
%if Eq.\ref{eq:pe-unified} has a unique solution $M^*$. The PE-GNN layer in Eq.~\ref{eq:pe-unified} is PE-stable with $C$, %Eq.\ref{eq:pe-unified} has a unique solution $M^*$; 2) PE-stable, if for two graphs $A^{(1)},\,A^{(2)}$, their unique solutions of Eq.\ref{eq:pe-unified} satisfies $\|M^*(A^{(1)}) - M^*(A^{(2)})\|_F \leq \|A^{(1)} - A^{(2)}\|$.
\end{theorem}

Note that the conditions on $f_+$ and $f_-$ are generally satisfied by Deepwalk, Node2vec and LINE. However, solving Eq.\ref{eq:pe-unified} to get the optimal solution may not be guaranteed in practice because the low-rank constraint is non-convex. One may consider relaxing the low-rank constraint into the nuclear norm $\|M\|_{*} \leq \tau$ for some threshold $\tau$~\citep{recht2010guaranteed}, which reduces the optimization Eq.\ref{eq:pe-unified} into a convex optimization and thus satisfies the conditions in Thm.~\ref{thm:general-pe}. Empirically, this step and the step of computing SVD seem unnecessary according to our experiments. The PE-stability is related to the value of the smallest non-zero singular value of $M^*$. We leave the full characterization of PE-equivariance and PE-stability for the general PE techniques for the future study.

\section{Experiments} \label{sec:exp}
%\vspace{-3mm}
In this work, we use the most important node-set-based task link prediction to evaluate \proj, though it may apply to more general tasks. %We do not consider node classification tasks because positional features are usually not relevant in these tasks. 
Two types of link prediction tasks are investigated: traditional link prediction (Task 1) and domain-shift link prediction (Task 2). In Task 1, the model gets trained, validated and tested over the same graphs while using different link sets. In Task 2, the graph used for training/validation is different from the one used for testing. Both tasks may reflect the effectiveness of a model while Task 2 may better demonstrate the model's generalization capability that strongly depends on permutation equivariance and stability. All the results are based on 10 times random tests.

%The details of the datasets and the experimental settings are introduced in Appendix.

%\vspace{-2mm}
\subsection{The Experimental Pipeline of \proj  for Link Prediction}
%\vspace{-2mm}
 We use \proj to build GNNs. The pipeline contains three steps. First, we adopt certain PE techniques to compute positional features $Z$. Second, we stack \proj layers according to Eq.~\ref{eq:PE-GNN-layer1}. Suppose the final node representations are denoted by $(\hat{X},\,Z)$. Third, for link prediction over  $(u,v)$, we concatenate $(\hat{X}_u\hat{X}_v^T,\,Z_uZ_v^T)$ denoted as $H_{uv}$ and adopt MLP$(H_{uv})$ to make final predictions. In the experiments, we test LE and Deepwalk as PE, and name the models \proj-LE and \proj-DW respectively. To verify the wide applicability of our theory, we also apply \proj to GraphSAGE~\cite{hamilton2017inductive} by similarly using the distance between PEs to control the neighbor aggregation according to Eq.~\ref{eq:PE-GNN-layer1} and term the corresponding models PE-SAGE-LE and PE-SAGE-DW respectively.

For some graphs, especially those small ones where the union of link sets for training, validation and testing cover the entire graph, the model may overfit positional features and hold a large generalization gap. This is because the links used as labels to supervise the model training are also used to generate positional features. To avoid this issue, we consider a more elegant way to use the training links. We adopt a 10-fold partition of the training set. For each epoch, we periodically pick one fold of the links to supervise the model while using the rest links to compute positional features. Note that PEs for different folds can be pre-computed by removing every fold of links, which reduces computational overhead. In practice, the testing stage often corresponds to online service and has stricter time constraints. Such partition is not needed so there is no computational overhead for testing. We term the models trained in this way as \proj-LE+ and \proj-DW+ respectively. %To verify the wide applicability of our theory on PEs, we also apply \proj to GraphSAGE when performing aggregation from neighbors and term the model                
%\vspace{-3mm}
\subsection{Task 1 --- Traditional link prediction}
%\vspace{-2mm}
\textbf{Datasets.}
We use eight real graphs, including Cora, CiteSeer and Pubmed \citep{sen2008collective}, Twitch (RU), Twitch (PT) and Chameleon~\citep{rozemberczki2021multi}, DDI and COLLAB \citep{hu2021ogb}. Over the first 6 graphs, we utilize 85\%, 5\%, 10\% to partition the link set that gives positive examples for training, validation and testing and pair them with the same numbers of negative examples (missing edges). For the two OGB graphs, we adopt the dataset splits in~\citep{hu2020open}. The links for validation and test are removed during the training stage and the links for validation are also not used in the test stage. All the models are trained till the loss converges and the models with the best validation performance is used to test.

 %GAE utilizes GCN to encode both the structural information and node feature information, and then decode the node embeddings to reconstruct the graph adjacency matrix. 
    %Variational Graph Autoencoder (VGAE) is a variational version of GAE, which can overcome the overfitting problem caused by the capacity of autoencoders.
    
\textbf{Baselines.}
    We choose 6 baselines: \emph{VGAE} \citep{kipf2016variational}, \emph{P-GNN} \citep{you2019position}, \emph{SEAL} \citep{zhang2018link}, \emph{GNN Trans.} \citep{dwivedi2020generalization}, \emph{LE}~\citep{belkin2003laplacian} and \emph{Deepwalk (DW)}~\citep{perozzi2014deepwalk}.
    VGAE is a variational version of GAE~\citep{kipf2016variational} that utilizes GCN to encode both the structural information and node feature information, and then decode the node embeddings to reconstruct the graph adjacency matrix. %The original VGAE does not use any distance feautures. 
    SEAL is particularly designed for link prediction by using enclosing subgraph representations of target node-pairs. %and it outperforms many other link prediction methods such as VGAE.
    %SEAL is a representative that uses DE.
    P-GNN randomly chooses some anchor nodes and aggregates only from these anchor nodes.  GNN Trans. adopts LE as PE and merges LE into node features and utilizes attention mechanism to aggregate information from neighbor nodes. For VGAE, P-GNN and GNN Trans., the inner product of two node embeddings is adopted to represent links.
    LE and DW are two network embedding methods, where the obtained node embeddings are already positional features and directly used to predict links.
    
    %To get a full comparison, we consider pairing different methods with different features. We consider 7 types of features: (1) original node features (N.); (2) node degree features (P.); (3) positional feature (P.) extracted by Deepwalk; (3) positional features (L.) extracted by LE; (5) one-hot feature (O.): one-hot encoding of node indices;  (6) random feature (R.); (7) distance features according to the target links (D.) that is only used by SEAL. 

    %Experiments are repeated 10 times and we report the average.
    
    %the baseline VGAE with different features: (1) node feature (N.): original feature of each node. (2) constant feature (P.): node degree. (3) positional feature (P.): PE extracted by Deepwalk. (4) one-hot feature (O.): one-hot encoding of node indices.  (5) random feature (R.): random value $r_v \sim \text{Unif}(D)$. (6) node feature and positional feature (N. + P.): concatenating the node feature and the positional feature.  
\textbf{Implementation details.} 
For VGAE, we consider six types of features: (1) node feature (N.): original feature of each node. (2) constant feature (C.): node degree. (3) positional feature (P.): PE extracted by Deepwalk. (4) one-hot feature (O.): one-hot encoding of node indices.  (5) random feature (R.): random value $r_v \sim U(0,1)$ (6) node feature and positional feature (N. + P.): concatenating the node feature and the positional feature. P-GNN uses node features and the distances from a node to some randomly selected anchor nodes as positional features (N. + P.). GNN Trans. utilizes node features and LE as positional features (N. + P.). SEAL adopt Double-Radius Node Labeling (DRNL) to compute deterministic distance features (N. + D.). For \proj, we consider node features plus positional features (N. + P.) or constant feature  plus positional features (C. + P.). 

%to reflect nodes’ relative positions and structural importance within subgraphs (N. + D.). For PEG, we consider node feature (N. + P.) and constant feature (C. + P.). 

\begin{table}[t]
%\vspace{-0mm}
\caption{Performance on the traditional link prediction tasks, measured in ROC AUC (mean$\pm$std\%).} %Mean $\pm$ Standard deviations are given. \pan{We need VGAE with node feature + positional feature}. }
%\vspace{-2mm}
\label{table1}
\centering
\resizebox{0.96\textwidth}{!}{
\begin{tabular}{c|c|c|c|c|c|c|c}
\hline
\multicolumn{1}{l|}{Method} & \multicolumn{1}{l|}{Feature} & Cora         & Citeseer     & Pubmed       & Twitch-RU    & Twitch-PT    & Chameleon    \\ \hline
\multirow{6}{*}{VGAE}       & N.                           & 89.89 ± 0.06 & 90.11 ± 0.08 & 94.62 ± 0.02 & 83.13 ± 0.07 & 82.89 ± 0.08 & 97.98 ± 0.01 \\
                            & C.                           & 55.68 ± 0.05 & 61.45 ± 0.36 & 69.03 ± 0.03 & 85.37 ± 0.02 & 85.69 ± 0.09 & 83.13 ± 0.04 \\
                            & O.                           & 83.97 ± 0.05 & 77.22 ± 0.04 & 82.54 ± 0.04 & 84.76 ± 0.09 & 87.91 ± 0.05 & 97.67 ± 0.04 \\
                            & P.                           & 83.82 ± 0.12 & 78.68 ± 0.25 & 81.74 ± 0.15 & 85.06 ± 0.14 & 85.06 ± 0.14 & 97.91 ± 0.03 \\
                            & R.                           & 68.43 ± 0.42 & 71.21 ± 0.78 & 69.31 ± 0.23 & 68.42 ± 0.43 & 68.49 ± 0.73 & 73.44 ± 0.53 \\
                            & \multicolumn{1}{l|}{N. + P.} & 87.96 ± 0.29 & 80.04 ± 0.60 & 85.26 ± 0.17 & 84.59 ± 0.37 & 88.27 ± 0.19 & 98.01 ± 0.12 \\ \hline
PGNN                        & N. + P.                      & 86.92 ± 0.02 & 90.26 ± 0.02 & 88.12 ± 0.06 & 83.21 ± 0.00 & 82.37 ± 0.02 & 94.25 ± 0.01 \\
GNN-Trans.                  & N. + P.                      & 79.31 ± 0.09 & 77.49 ± 0.02 & 81.23 ± 0.12 & 79.24 ± 0.33 & 75.44 ± 0.14 & 86.23 ± 0.12 \\
SEAL                        & N. + D.                      & 91.32 ± 0.91 & 89.49 ± 0.43 & 97.16 ± 0.28 & 92.12 ± 0.10 & \textbf{93.21 ± 0.06}$^\dagger$ & \textbf{99.31 ± 0.18}$^\dagger$ \\ \hline
LE                          & P.                           & 84.43 ± 0.02 & 78.36 ± 0.08 & 84.35 ± 0.04 & 78.80 ± 0.10 & 67.56 ± 0.02 & 88.47 ± 0.03 \\
DW                          & P.                           & 86.82 ± 0.18 & 87.93 ± 0.11 & 85.79 ± 0.06 & 83.10 ± 0.05 & 83.47 ± 0.03 & 92.15 ± 0.02 \\ \hline
PEG-DW                      & N. + P.                      & 89.51 ± 0.08 & 91.67 ± 0.12 & 87.68 ± 0.29 & 90.21 ± 0.04 & 89.67 ± 0.03 & 98.33 ± 0.01 \\
PEG-DW                      & C. + P.                      & 88.36 ± 0.10 & 88.48 ± 0.10 & 88.80 ± 0.11 & 90.32 ± 0.09 & 90.88 ± 0.05 & 97.30 ± 0.03 \\
PEG-LE                      & N. + P.                      & \textbf{94.20 ± 0.04}$^\dagger$ & 92.53 ± 0.09 & 87.70 ± 0.31 & 92.14 ± 0.05 & 92.28 ± 0.02 & 98.78 ± 0.02 \\
PEG-LE                      & C. + P.                      & 86.88 ± 0.03 & 76.96 ± 0.23 & 91.65 ± 0.02 & 90.21 ± 0.18 & 91.15 ± 0.13 & 98.73 ± 0.04 \\ \hline
PEG-DW+                     & N. + P.                      & 93.32 ± 0.08 & 94.11 ± 0.14 & 97.88 ± 0.05 & 91.68 ± 0.01 & 92.15 ± 0.02 & 98.20 ± 0.01 \\
PEG-DW+                     & C. + P.                      & 90.78 ± 0.09 & 91.22 ± 0.12 & 93.44 ± 0.05 & 90.22 ± 0.04 & 91.37 ± 0.05 & 97.50 ± 0.03 \\
PEG-LE+                     & N. + P.                      & 93.78 ± 0.03 & \textbf{95.73 ± 0.09}$^\dagger$ & \textbf{97.92 ± 0.11}$^\dagger$ & 92.29 ± 0.11 & 92.37 ± 0.06 & 98.18 ± 0.02 \\
PEG-LE+                     & C. + P.                      & 88.98 ± 0.14 & 78.61 ± 0.27 & 94.28 ± 0.05 & \textbf{92.35 ± 0.02}$^\dagger$ & 92.50 ± 0.06 & 97.79 ± 0.01 \\ \hline
\end{tabular}}
%\vspace{-3mm}
\end{table}

\begin{table}[t]
\caption{Performance on OGB datasets, measured in Hit@20 and Hits@50 (mean$\pm$std\%). Codes are run on CPU: Intel(R) Xeon(R) Gold 6248R @ 3.00GHz and GPU: NVIDIA QUADRO RTX 6000.}
%\vspace{-2mm}
\scriptsize
\label{table2}
\resizebox{\textwidth}{!}{
\begin{tabular}{c|cccc|cccc}
\hline
\multirow{2}{*}{Method} & \multicolumn{4}{c|}{ogbl-ddi (Hits@20(\%))}                           & \multicolumn{4}{c}{ogbl-collab (Hits@50(\%))}                         \\ \cline{2-9} 
                        %& \multicolumn{4}{c|}{Hits@20(\%)}                        & \multicolumn{4}{c}{Hits@50(\%)}                         \\
                        & training time & test time & Validation   & test         & training time & test time & Validation   & test         \\ \hline
GCN                     & 29min 27s     & 0.20s     & 55.27 ± 0.53 & 37.11 ± 0.21 & 1h38min17s    & 1.38s     & 52.71 ± 0.10 & 44.62 ± 0.01 \\
GraphSAGE                    & 14min 26s  & 0.24s    & 67.11 ± 1.21 & 52.81 ± 8.75 & 38min 10s    & 0.83s    & 57.16 ± 0.70 & 48.45 ± 0.80 \\
SEAL                    & 2h 04min 32s  & 12.04s    & 28.29 ± 0.38 & 30.23 ± 0.24 & 2h29min05s    & 51.28s    & 64.95 ± 0.04 & \textbf{54.71 ± 0.01}$^\dagger$ \\
PGNN                    & 9min 49.39s   & 0.28s     & 2.66 ± 0.16  & 1.74 ± 0.19  & N/A           & N/A       & N/A          & N/A          \\
GNN-trans.              & 53min 26s     & 0.35s     & 15.63 ± 0.14 & 9.22 ± 0.21  & 1h52min22s    & 1.86s     & 18.17 ± 0.25 & 11.19 ± 0.42 \\ \hline
DW                      & 36min 41s     & 0.23s     & 0.04 ± 0.00  & 0.02 ± 0.00  & 34min40s      & 1.08s     & 53.64 ± 0.03 & 44.79 ± 0.02 \\
LE                      & 33min 42s     & 0.29s     & 0.09 ± 0.00  & 0.02 ± 0.00  & 37min22s      & 1.23s     & 0.10 ± 0.01  & 0.12 ± 0.02  \\ \hline
PEG-DW                  & 29min 56s     & 0.27s     & 56.47 ± 0.35 & 43.80 ± 0.32 & 1h42min 05s   & 1.51s     & 63.98 ± 0.05 & 54.33 ± 0.06 \\
PEG-LE                  & 30min 32s     & 0.29s     & 57.49 ± 0.47 & 30.16 ± 0.47 & 1h42min03s    & 1.42s     & 56.52 ± 0.12 & 48.76 ± 0.92 \\
PE-SAGE-DW                  & 25min 11s  & 0.31s    & 68.05 ± 0.96 & \textbf{56.16 ± 5.50}$^\dagger$ & 56min54s    & 0.97s    & 63.43 ± 0.48 & 54.17 ± 0.54
 \\
PE-SAGE-LE                  & 26min 19s  & 0.32s    & 68.38 ± 0.78 & 51.49 ± 9.71 & 55min59s    & 0.98s    & 58.66 ± 0.55 &
49.75 ± 0.67  \\
PEG-DW+                 & 48min 03s     & 0.28s     & 59.70 ± 6.87 & 47.93 ± 0.21 & 1h37min43s    & 1.43s     & 62.31 ± 0.19 & 53.71 ± 8.02 \\
PEG-LE+                 & 51min 25.35s  & 0.29s     & 58.44 ± 1.71 & 28.32 ± 7.34 & 1h33min29s    & 1.39s     & 52.91 ± 1.24 & 45.96 ± 9.98 \\ \hline
\end{tabular}}
%\vspace{-5mm}
\end{table}

\textbf{Results} are shown in Table~\ref{table1} and Table~\ref{table2}. Over the small datasets in Table~\ref{table1}, VGAE with node features outperforms other features in Cora, Citeseer and Pubmed because the nodes features therein are mostly informative, while this is not true over the other three datasets. One-hot features and positional features almost achieve the same performance, which implies that that GNNs naively using PE makes positional features behave like one-hot features and may have instability issues. Constant features are not good because of the node ambiguity issues. Random features may introduce heavy noise that causes trouble in model convergence. Concatenating node features and positional features gives better performance than only using positional features but is sometimes worse than only using node features, which is again due to the instability issue by using positional features.      

%outperforms all kinds of other features since node features contains useful information. Constant features and random features perform bad because of their limited relationship with nodes. One-hot feature can distinguish nodes and positional feature can provide structural information, thus their performance is not bad on traditional link prediction task. Concatenating node feature and positional feature might be helpful sometimes, but not stable. 
Although PGNN and GNN Trans. utilize positional features, they achieve subpar performance. SEAL outperforms all of the state-of-art methods, which again demonstrates that the effectiveness of distance features~\citep{li2020distance,zhang2020revisiting}. 
%learning representations from links is better than node embedding combination in most link prediction cases. 
\proj significantly outperforms all the baselines except SEAL. PEG+ by better using training sets achieves comparable or even better performance than SEAL, which demonstrates the contributions of the stable usage of PE. Moreover, \proj can achieve comparable performance in most cases even only paired with constant features, which benefits from the more expressive power given by PE (avoids node ambiguity). Note that PE without GNNs (LE or DW solo) does not perform well, which justifies the benefit by joining GNN with PE. 

%In addition, DW seems to be better to be chosen as PE than LE, since it is more stable.

% can contribute to more powerful GNN. demonstrates that the

% \proj can achieve comparable performance with SEAL, and sometimes better than SEAL, which supports the fact that equivariant and stable positional encoding can contribute to more powerful GNN. Moreover, PE-GNN can achieve comparable performance in most cases when we only have constant feature. In addition, DW seems to be better to be chosen as PE than LE, since it is more stable.

Regarding the OGB datasets, PGNN and GNN Trans. do not perform well either. Besides, PGNN cannot scale to large graphs such as \emph{collab}. %, since it need to compute all-pairs shortest path distances of the network. 
The results of DW and LE demonstrate that the original positional features may only provide crude information, so pairing them with GNNs is helpful. % so when we use ranking evaluation metrics such as Hits, they don't work. 
\proj achieves the best results on \emph{ddi}, and performs competitively with SEAL on \emph{collab}. %Note SEAL has less precomputing time, since it only trains on 1\% training edge for ddi and 15\% for collab. %We extract DW PE by using OpenNE \footnote{https://github.com/thunlp/OpenNE/tree/pytorch}, which do all the computation on CPU, thus our DW is time consuming. For PEG-DW+, multi-process can be adopted to accelerate the computation, but in this experiment we use single thread because of the time limitation. 
The complexity of \proj is comparable to canonical GNNs. Note that we do not count the time of PE as it relates to the particular implementation. If \proj is used for online serving when the time complexity is more important, PE can be computed in prior. For a fair comparison, we also do not count the pre-processing time of SEAL, PGNN or GNN Trans.. Most importantly, \proj performs significantly faster than SEAL on test because SEAL needs to compute distance features for every link while PE in \proj is shared by links. %which is equally worth discussing since we train a model only once but we might test it many times. 
Interestingly, DW seems better than LE as a PE technique for large networks. %PEG-LE dos not perform well on ddi. One possible reason is that ddi is considerably denser than the other graphs, matrix decomposition might be hard to learn any meaningful structural patterns.

%%\vspace{-2mm}
\subsection{Task 2 ---Domain-shift link prediction}
%\vspace{-2mm}
\textbf{Datasets \& Baselines.} Task 2 better evaluates the generalization capability of models. We consider 3 groups, including citation networks (cora$\rightarrow$citeseer and cora$\rightarrow$pubmed) \citep{sen2008collective}, user-interaction networks (Twitch (EN)$\rightarrow$Twitch (RU) and Twitch (ES)$\rightarrow$Twitch (PT)) \citep{rozemberczki2021multi} and biological networks (PPI) \citep{hamilton2017inductive}. For citation networks and user-interaction networks, we utilize 95\%, 5\% dataset splitting for training and validation over the training graph, and we use 10\% existing links as test positive links in the test graph. For PPI dataset, we randomly select 3 graphs as training, validation and testing datasets and we sample 10\% existing links in the validation/testing graphs as validation/test positive links. %\vspace{-1mm}

%\textbf{Baselines.} We use the same baselines as transductive link prediction experiment.

\textbf{Baselines \& Implementation details.} We use the same baselines as Task 1, while we do not use one-hot features for VGAE, since the training and test graphs have different sizes. As for node features, we randomly project them into the same dimension and then perform row normalization on them. Other settings are the same as Task 1. PE is applied to training and testing graphs separately while PE over testing graphs is computed after removing the testing links. %Experiments are repeated 10 times and we report the average.
%\vspace{-1mm}

\begin{table}[t]
\caption{Performance on the domain-shift link prediction tasks, measured in ROC AUC (mean$\pm$std\%)}
%\vspace{-2mm}
\label{table3}
\centering
\resizebox{0.90\textwidth}{!}{
\begin{tabular}{c|c|c|c|c|c|c}
\hline
Mehood                 & Features & Cora$\rightarrow$Citeseer & Cora$\rightarrow$Pubmed  & ES$\rightarrow$PT & EN$\rightarrow$RU               & PPI          \\ \hline
                       & N.       & 62.74 ± 0.03  & 63.53 ± 0.27 & 51.52 ± 0.17          & 60.08 ± 0.02                        & 83.24 ± 0.20 \\
                       & C.       & 62.16 ± 0.08  & 56.89 ± 0.36 & 82.72 ± 0.26          & {\color[HTML]{000000} 91.23 ± 0.07} & 75.27 ± 0.89 \\
                       & P.       & 70.59 ± 0.03  & 79.83 ± 0.27 & 82.24 ± 0.24          & 81.42 ± 0.01                        & 77.61 ± 0.47 \\
                       & R.       & 68.44 ± 0.63  & 71.27 ± 0.37 & 71.26 ± 0.36          & 69.37 ± 0.35                        & 75.88 ± 0.49 \\
\multirow{-5}{*}{VGAE} & N. + P.  & 76.45 ± 0.55  & 65.62 ± 0.42 & 71.46 ± 0.31          & 84.00 ± 0.28                        & 84.67 ± 0.22 \\ \hline
PGNN                   & N. + P.  & 85.02 ± 0.28  & 76.88 ± 0.42 & 70.41 ± 0.07          & 63.27 ± 0.27                        & 80.84 ± 0.03 \\
GNN-Trans.             & N. + P.  & 61.60 ± 0.52  & 76.35 ± 0.17 & 63.44 ± 0.34          & 62.87 ± 0.22                        & 79.82 ± 0.17 \\
SEAL                   & N. + D.  & \textbf{91.36 ± 0.93}$^\dagger$  & 89.62 ± 0.87 & \textbf{93.37 ± 0.05}$^\dagger$          & 92.34 ± 0.14                        & 88.99 ± 0.12 \\ \hline
LE                     & P.       & 77.62 ± 0.04  & 84.03 ± 0.22 & 67.75 ± 0.09          & 77.57 ± 0.15                        & 72.14 ± 0.82 \\
DW                     & P.       & 86.48 ± 0.14  & 86.97 ± 0.06 & 83.56 ± 0.03          & 83.41 ± 0.04                        & 85.18 ± 0.20 \\ \hline
PEG-DW                 & N. + P.  & 89.91 ± 0.03  & 87.23 ± 0.34 & 91.82 ± 0.04          & 91.14 ± 0.02                        & 87.36 ± 0.11 \\
PEG-DW                 & C. + P.  & 89.75 ± 0.04  & 89.58 ± 0.08 & 91.27 ± 0.04          & 90.26 ± 0.07                        & 86.42 ± 0.20 \\
PEG-LE                 & N. + P.  & 82.57 ± 0.02  & 92.34 ± 0.28 & 91.61 ± 0.05          & 91.93 ± 0.13                        & 85.34 ± 0.14 \\
PEG-LE                 & C. + P.  & 79.60 ± 0.04  & 88.89 ± 0.13 & 91.38 ± 0.10          & 92.40 ± 0.10                        & 85.22 ± 0.16 \\ \hline
PEG-DW+                & N. + P.  & 91.15 ± 0.06  & 90.98 ± 0.03 & 91.24 ± 0.16          & 91.91 ± 0.02                        & \textbf{89.92 ± 0.17}$^\dagger$ \\
PEG-DW+                & C. + P.  & 91.32 ± 0.01  & 90.93 ± 0.18 & 91.22 ± 0.02          & 92.14 ± 0.02                        & 88.44 ± 0.29 \\
PEG-LE+                & N. + P.  & 86.72 ± 0.05  & \textbf{93.34 ± 0.11}$^\dagger$ & 91.67 ± 0.13          & 92.24 ± 0.19                        & 86.77 ± 0.36 \\
PEG-LE+                & C. + P.  & 87.62 ± 0.04  & 92.21 ± 0.20 & 91.37 ± 0.19          & \textbf{93.12 ± 0.21}$^\dagger$                        & 86.21 ± 0.27 \\ \hline
\end{tabular}}
%\vspace{-5mm}
\end{table}

\textbf{Results} are shown in Table~\ref{table3}. Compared with Table~\ref{table1}, for VGAE, we notice that node features perform much worse (except PPI) than Task 1, which demonstrates the risks of using node features when the domain shifts. Positional features, which is not specified for the same graph, is possibly more generalizable over different graphs. Random features are generalizable while still hard to converge. %, and it achieves subpar performance. 
PGNN and GNN Trans. do not utilize positional features appropriately and perform far from ideal. %thus they have low generalization capacity. 
Both SEAL and \proj outperform other baselines significantly, which implies their good stability and generalization. \proj and SEAL again achieve comparable performance while \proj has much lower training and testing complexity. Our results successfully demonstrate the significance of using permutation equivariant and stable PE.

%\vspace{-3mm}
\section{Conclusion}
%\vspace{-2mm}
In this work, we studied how GNNs should work with PE in principle, and proposed the conditions that keep GNNs permutation equivariant and stable when PE is used to avoid the node ambiguity issue. We follow those conditions and propose the \proj layer. Extensive experiments on link prediction demonstrate the effectiveness of \proj. In the future, we plan to generalize the theory to more general PE techniques and test \proj over other graph learning tasks.

\subsubsection*{Acknowledgments}
We greatly thank the actionable suggestions given by reviewers. H. Yin and P. L. are supported by the 2021 JPMorgan Faculty Award and the National Science Foundation (NSF) award HDR-2117997.

\bibliography{BIB/gnn-position,BIB/gnn-background,BIB/general-background,BIB/network-embedding}
\bibliographystyle{iclr2022_conference}

\appendix

\section{Proof of Lemma~\ref{thm:unstable}}
Recall the eigenvalues of the PSD matrix $B$ are $0=\lambda_1<\lambda_2 <...< \lambda_{p}< \lambda_{p+1} \leq \lambda_{p+2} \leq ... \leq \lambda_{N}$. %\htyin{$\le  \lambda_{p+2}$?} 
Suppose one EVD of $B = U\Lambda U^T$. In $U=[u_1,u_2,...,u_N]$, $u_i$ is the eigenvector of $\lambda_i$ so $Bu_{i} = \lambda_i u_i$. Without loss of generality, we set $\text{PE}(B) = [u_1,u_2,...,u_p]$.

Suppose $k=\argmin_{1\leq i\leq p}|\lambda_{i+1} - \lambda_i|$. Now, we perturb $B$ by slightly  perturbing $u_k$ and $u_{k+1}$. We set 
\begin{align*}
    u_{k}' &= \sqrt{1-\epsilon^2} u_k + \epsilon u_{k+1} \\
    u_{k+1}' &= - \epsilon u_{k} + \sqrt{1-\epsilon^2} u_{k+1} 
\end{align*}
Set $u_{i}' = u_{i}$ for $i\in[N], i\neq k,k+1$. Note that $\|u_{k}'\|=\|u_{k+1}'\|=1$ and $u_{k}'^Tu_{k+1}'=0$. Then, the columns of $U'=[u_1',u_2',...,u_{N}']$ still give a group of orthogonal bases. 

Now we denote the above perturbation of $B$ as $B+\triangle B = \sum_{i=1}^{N} \lambda_{i}u_i'u_i'^{T}$. Then, $\text{PE}(B+\triangle B)$ could be $[u_1',u_2',...,u_p']S'$ for any $S'\in \text{SN}(p)$. Therefore, for sufficiently small $\epsilon>0$,
\begin{align}
    &\min_{S'\in \text{SN}(p)}\|\text{PE}(B+\triangle B)S' -  \text{PE}(B)\|_{\text{F}}^2 \nonumber \\
    =& \|[ (\sqrt{1-\epsilon^2}-1)u_k+ \epsilon u_{k+1},  (\sqrt{1-\epsilon^2}-1)u_{k+1}- \epsilon u_{k}]\|_{\text{F}}^2\nonumber\\
    =& \|(\sqrt{1-\epsilon^2}-1)u_k+ \epsilon u_{k+1}\|^2 + \|(\sqrt{1-\epsilon^2}-1)u_{k+1}- \epsilon u_{k}\|^2\nonumber \\
    =& 4(1- \sqrt{1-\epsilon^2}) \nonumber\\
    =& 2\epsilon^2 + o(\epsilon^2).  \label{apd:proofB11}
\end{align}

Next, we characterize $\|\triangle B\|_{\text{F}}$. 
\begin{align}
    &\|\triangle B\|^2_{\text{F}}  =\|B+\triangle B - B\|^2_{\text{F}} = \|\sum_{i=1}^{N} \lambda_{i}u_i'u_i'^{T} - \sum_{i=1}^{N} \lambda_{i}u_iu_i^{T}\|^2_{\text{F}} \nonumber \\
    =& \| \lambda_k (u_{k}'u_{k}'^T - u_{k}u_{k}^T) + \lambda_{k+1}(u_{k+1}'u_{k+1}'^T - u_{k+1}u_{k+1}^T) \|^2_{\text{F}}\nonumber \\
    =& \| (\lambda_{k+1}-\lambda_{k}) \left[-\epsilon^2(u_{k}u_{k}^T - u_{k+1}u_{k+1}^T) + \epsilon\sqrt{1-\epsilon^2}(u_{k}u_{k+1}^T + u_{k+1}u_{k}^T) \right] \|^2_{\text{F}} \nonumber \\
    =&(\lambda_{k+1}-\lambda_{k})^2( \epsilon^2 \|u_{k}u_{k+1}^T + u_{k+1}u_{k}^T\|^2_{\text{F}} + o(\epsilon^2)) \nonumber \\
        =&2(\lambda_{k+1}-\lambda_{k})^2 (\epsilon^2  + o(\epsilon^2)). \label{apd:proofB12}
\end{align}
Combining Eqs.~\ref{apd:proofB11},\ref{apd:proofB12}, we have, for sufficiently small $\epsilon>0$,
\begin{align*}
\min_{S'\in \text{SN}(p)}\|\text{PE}(B+\triangle B)S' -  \text{PE}(B)\|_{\text{F}} > 0.99|\lambda_{k+1}-\lambda_{k}|^{-1}\|\triangle B\|_{\text{F}} + o(\epsilon),
\end{align*}
which concludes the proof.

\section{Proof of Lemma~\ref{thm:stable}}

The result of the Lemma~\ref{thm:stable} can be derived from the Davis-Kahan theorem~\citep{davis1970rotation} and its variant~\citep{yu2015useful} that characterizes the eigenspace perturbation. We apply the Theorem 2 Eq.3 of~\citep{yu2015useful} to two PSD matrices.  

\begin{theorem}[Theorem 2~\citep{yu2015useful}]
Let two PSD matrices $B^{(1)},\,B^{(2)}\in \mathbb{R}^{N\times N}$, with eigenvalues $0=\lambda_1^{(i)}\leq \lambda_2^{(i)}\leq...\leq\lambda_p^{(i)}\leq ...\leq \lambda_p^{(N)}$ such that $\lambda_{p+1}^{(1)} - \lambda_{p}^{(1)}>0$. For $i=1,2$, let $U^{(i)} = (u_1^{(i)}, u_2^{(i)},...,u_p^{(i)})$ have orthonormal columns satisfying $B^{(i)}u_k^{(i)} = \lambda_ku_k^{(i)}$ for $k=1,2,...,p$. Then, there exists an orthogonal matrix $Q\in\text{O}(p)$ such that 
\begin{align*}
\|U^{(2)}Q - U^{(1)}\|_\text{F} \leq \frac{2^{3/2}\min(p^{1/2}\|B^{(1)}-B^{(2)}\|_\text{op},\|B^{(1)}-B^{(2)}\|_\text{F})}{\lambda_{p+1}^{(1)} - \lambda_{p}^{(1)}}
\end{align*}
\end{theorem}

By symmetry, use the above theorem again and we know there exists $Q'\in\text{O}(p)$
\begin{align*}
\|U^{(1)}Q' - U^{(2)}\|_\text{F} \leq \frac{2^{3/2}\min(p^{1/2}\|B^{(1)}-B^{(2)}\|_\text{op},\|B^{(1)}-B^{(2)}\|_\text{F})}{\lambda_{p+1}^{(2)} - \lambda_{p}^{(2)}}
\end{align*}

Because $\|U^{(1)}Q' - U^{(2)}\|_\text{F} = \|U^{(2)}Q'^T - U^{(1)}\|_\text{F}$, then there exists $Q\in\text{O}(p)$, 
\begin{align} \label{adp:proofC11}
\|U^{(2)}Q - U^{(1)}\|_\text{F} \leq 2^{3/2}\delta\min(p^{1/2}\|B^{(1)}-B^{(2)}\|_\text{op},\|B^{(1)}-B^{(2)}\|_\text{F})
\end{align}
where $\delta = \min\{(\lambda_{p+1}^{(1)} - \lambda_{p}^{(1)})^{-1}, (\lambda_{p+1}^{(2)} - \lambda_{p}^{(2)})^{-1}\}$.

When we apply a permutation matrix $P\in \Pi$ to permute the rows and columns of $B^{(2)}$, then $(PB^{(2)}P^T)(Pu_{k}^{(2)}) = PB^{(2)}u_{k}^{(2)}=\lambda_k^{(2)}Pu_{k}^{(2)}$ for any $k$. Moreover, permuting the rows and columns of a PSD matrix will not change its eigenvalues. This means that $B^{(2)}$ can be replaced by $PB^{(2)}P^T$ in Eq.\ref{adp:proofC11} as long as $U^{(2)}$ is replaced by $PU^{(2)}$. Therefore, 

\begin{align*}
\|PU^{(2)}Q - U^{(1)}\|_\text{F} \leq 2^{3/2}\delta\min(p^{1/2}\|B^{(1)}-PB^{(2)}P^T\|_\text{op},\|B^{(1)}-PB^{(2)}P^T\|_\text{F})
\end{align*}

%By further simply setting $b=\mathbf{0}$, we prove Lemma~\ref{thm:stable}.

%and there exists a $b\in\mathbb{R}^{p}$ su %\sqrt{tr\left((U^{(1)}Q' - U^{(2)})(U^{(1)}Q' - U^{(2)})^T\right)}= \sqrt{tr\left((U^{(1)} - U^{(2)}Q'^T)(U^{(1)} - U^{(2)}Q'^T)^T\right)}=\|U^{(2)}Q'^T - U^{(1)}\|_\text{F}$

\section{Proof of Theorem~\ref{thm:equivariant}}
To prove PE-equivariance, consider two graphs $\gG^{(1)} = (A^{(1)}, X^{(1)})$ and  $\gG^{(2)} = (A^{(2)}, X^{(2)})$ that have perfect matching $P^*$. So, $L^{(1)} = P^*L^{(2)}P^{*T}$, $X^{(1)} = P^*X^{(2)}$. 

Let $Z^{(i)}$ denote the Laplacian eigenmaps of $L^{(i)}$, $i=1,2$. Set $B^{(i)} = L^{(i)}$, $i=1,2$ and $P=P^*$ and use Lemma~\ref{thm:stable}. Because $L^{(1)} = P^*L^{(2)}P^{*T}$ and $\lambda_{p}\neq \lambda_{p+1}$, then there exists $Q\in\text{O}(p)$. %and $b\in\mathbb{R}^p$.
\begin{align}\label{apd:proofD11}
    Z^{(1)} = P^*Z^{(2)}Q %+ \mathbf{1}b^T,
\end{align}
%where $Z^{(i)}$ is the Laplacian eigenmap of $L^{(i)}$. 

Now, we consider a GNN layer $g$ that satisfies Eqs.~\ref{eq:pi-eq},\ref{eq:oth-eq}. Also denote the output as $(\hat{X}^{(1)}, \hat{Z}^{(1)})=g(A^{(1)}, X^{(1)}, Z^{(1)})$ and $(\hat{X}^{(2)}, \hat{Z}^{(2)}) = g(A^{(2)}, X^{(2)}, Z^{(2)})$.
\begin{align*}
    (\hat{X}^{(1)}, \hat{Z}^{(1)}) & = g(A^{(1)}, X^{(1)}, Z^{(1)})  \\ 
    &\stackrel{(a)}{=} g(P^*A^{(2)}P^{*T}, P^{*T}X^{(2)},  P^*Z^{(2)}Q )\\
    &\stackrel{(b)}{=} P^*g(A^{(2)}, X^{(2)}, Z^{(2)}Q) \\
    &\stackrel{(c)}{=}  (P^*\hat{X}^{(2)}, P^*\hat{Z}^{(2)}Q)
\end{align*}
Here (a) is because the perfect matching between $\gG^{(1)}$ and  $\gG^{(2)}$, and Eq.~\ref{apd:proofD11}. (b) is due to Eq.~\ref{eq:pi-eq} and (c) is due to Eq.~\ref{eq:oth-eq}.

Therefore, $\hat{X}^{(1)} = P^*\hat{X}^{(2)}$ and $\eta( \hat{Z}^{(1)},  \hat{Z}^{(2)}) = 0$, which implies that $g$ satisfies PE-equivariance.

%If $(\hat{X}^{(1)}, \hat{Z}^{(1)})=g(A^{(1)}, X^{(1)}, Z^{(1)})$ and $(\hat{X}^{(2)}, \hat{Z}^{(2)}) = g(A^{(2)}, X^{(2)}, Z^{(2)})$, 

\section{Proof of Theorem~\ref{thm:lap}}
To prove PE-stability, consider two graphs $\gG^{(1)} = (A^{(1)}, X^{(1)})$ and  $\gG^{(2)} = (A^{(2)}, X^{(2)})$. Let $P^*$ denote their matching. We study the \proj layer in Eq.~\ref{eq:PE-GNN-layer1} and denote $(\hat{X}^{(i)}, \hat{Z}^{(i)}) = g_{\proj}(A^{(i)}, X^{(i)}, Z^{(i)})$ for $i=1,2$.

Let us first bound the easier term regarding the positional features $\eta(\hat{Z}^{(1)}, P^*\hat{Z}^{(2)})$. Because $\hat{Z}^{(i)}=Z^{(i)}$, $\eta(\hat{Z}^{(1)}, P^*\hat{Z}^{(2)}) = \eta(Z^{(1)}, P^*Z^{(2)})$, while the bound of the later is given by Lemma~\ref{thm:stable}. Set $B^{(i)} = L^{(i)}$, $i=1,2$ and $P=P^*$. Then, there exists a $Q\in\text{O}(p)$,  %Because $L^{(1)} = P^*L^{(2)}P^{*T}$ and $\lambda_{p}\neq \lambda_{p+1}$, then there exist $Q\in\text{O}(p)$.
\begin{align} \label{apd:proofE11}
    \|P^*Z^{(2)}Q - Z^{(1)}\|_\text{F} \leq 2^{3/2}\delta\min(p^{1/2}\|L^{(1)}-P^*L^{(2)}P^{*T}\|_\text{op},\|L^{(1)}-P^*L^{(2)}P^{*T}\|_\text{F}).
\end{align}

Next, we bound the harder part $\|\hat{X}^{(1)}- P^*\hat{X}^{(2)}\|_{\text{F}}$. First,
\begin{align} 
\|\hat{X}^{(1)}- P^*\hat{X}^{(2)}\|_{\text{F}}=\|\psi\left[\left(\hat{A}^{(1)} \odot \Xi^{(1)}\right) X^{(1)} W \right] - P^*\psi\left[\left(\hat{A}^{(2)} \odot \Xi^{(2)}\right) X^{(2)} W \right]\|_{\text{F}}.
\label{apd:proofE10}
\end{align}
Here, without loss of generality, we set $\Xi_{uv}^{(i)}= \phi(\|Z_u^{(i)}-Z_v^{(i)}\|)$ for $u,v\in[N]$ such that $\hat{A}^{(1)}_{uv}\neq 0$ and otherwise 0. Moreover, $\Xi_{uv}^{(i)}$ is bounded because $\Xi_{uv}^{(i)}= \phi(\|Z_u^{(i)}-Z_v^{(i)}\|) \leq \ell_{\phi}\|Z_u^{(i)}-Z_v^{(i)}\|\leq \ell_{\phi}(\|Z_u^{(i)}\|+\|Z_v^{(i)}\|)\leq 2\ell_{\phi}$ because of the $\ell_{\phi}$-Lipschitz continuity of $\phi$ and $\|Z_u^{(i)}\|\leq 1$.

Compute the difference  
\small{
\begin{align} 
&\|\psi\left[\left(\hat{A}^{(1)} \odot \Xi^{(1)}\right) X^{(1)} W \right] - P^*\psi\left[\left(\hat{A}^{(2)} \odot \Xi^{(2)}\right) X^{(2)} W \right]\|_{\text{F}}  \nonumber\\
&\stackrel{(a)}{\leq}\ell_\psi \|\left(\hat{A}^{(1)} \odot \Xi^{(1)}\right) X^{(1)} W - P^*\left(\hat{A}^{(2)} \odot \Xi^{(2)}\right) X^{(2)} W  \|_{\text{F}}  \nonumber\\
&\stackrel{(b)}{\leq} \ell_\psi \|W\|_{\text{op}} \|\left(\hat{A}^{(1)} \odot \Xi^{(1)}\right) X^{(1)} - P^*\left(\hat{A}^{(2)} \odot \Xi^{(2)}\right) X^{(2)} \|_{\text{F}}  \nonumber\\
& \stackrel{(c)}{=} \ell_\psi \|W\|_{\text{op}}  \|\left(\hat{A}^{(1)} \odot \Xi^{(1)}-P^*\left(\hat{A}^{(2)} \odot \Xi^{(2)}\right)P^{*T}\right) X^{(1)} +  P^{*}\left(\hat{A}^{(2)} \odot \Xi^{(2)}\right) (P^{*T}X^{(1)}-X^{(2)}) \|_{\text{F}}  \nonumber\\
%& \leq \ell_\psi \|W\|_{\text{op}} \left[\|\left(\hat{A}^{(1)} \odot \Xi^{(1)}-\hat{A}^{(2)} \odot \Xi^{(2)}\right) X^{(1)} \|_{\text{F}} + \|\left(\hat{A}^{(2)} \odot \Xi^{(2)}\right) (X^{(1)}-X^{(2)}) \|_{\text{F}}\right]  \nonumber \\
&\stackrel{(d)}{\leq} \ell_\psi \|W\|_{\text{op}}\left[\|\hat{A}^{(1)} \odot \Xi^{(1)}-P^{*}\left(\hat{A}^{(2)} \odot \Xi^{(2)}\right)P^{*T}\|_{\text{F}} \|X^{(1)} \|_{\text{op}} + \right. \\
&\quad\quad\quad\quad\quad\quad\quad\quad\quad\quad\quad\quad\quad\quad\quad\quad\quad\quad\quad\quad\quad\quad\quad\quad\left.\|P^{*T}\left(\hat{A}^{(2)} \odot \Xi^{(2)}\right)\|_{\text{op}} \|P^{*T}X^{(1)}-X^{(2)}\|_{\text{F}}\right] \nonumber\\
&\stackrel{(e)}{\leq} \ell_\psi \|W\|_{\text{op}}\left[\|\hat{A}^{(1)} \odot \Xi^{(1)}-P^{*}\left(\hat{A}^{(2)} \odot \Xi^{(2)}\right)P^{*T}\|_{\text{F}} \|X^{(1)} \|_{\text{op}} + \|\hat{A}^{(2)} \odot \Xi^{(2)}\|_{\text{op}} \|X^{(1)}-P^*X^{(2)}\|_{\text{F}}\right] 
\label{apd:proofE12}
\end{align}}
\normalsize
Here, (a) is due to the $\ell_\psi$-Lipschitz continuity of the component-wise function $\psi$. (b) and (d) use that for any two matrices $A,B$, $\|AB\|_{\text{F}}\leq \min\{\|A\|_{\text{F}}\|B\|_{\text{op}}, \|A\|_{\text{op}}\|B\|_{\text{F}}\}$ and (d) also uses the triangle inequality of $\|\cdot \|_{\text{F}}$. (e) uses $\|PA\|_{\text{op}} = \|A\|_{\text{op}}$ and $\|PA\|_{\text{F}} = \|A\|_{\text{F}}$ for any $P\in \Pi$.

Next, we only need to bound $\|\hat{A}^{(1)} \odot \Xi^{(1)}-P^{*}\left(\hat{A}^{(2)} \odot \Xi^{(2)}\right)P^{*T}\|_{\text{F}}$ and $\|\hat{A}^{(2)} \odot \Xi^{(2)}\|_{\text{op}}$. 
\begin{align} 
&\|\hat{A}^{(1)} \odot \Xi^{(1)}-P^{*}\left(\hat{A}^{(2)} \odot \Xi^{(2)}\right)P^{*T}\|_{\text{F}} \nonumber \\
& \stackrel{(a)}{=} \|\hat{A}^{(1)} \odot \Xi^{(1)}-\left(P^{*}\hat{A}^{(2)}P^{*T} \right)\odot \left(P^{*}\Xi^{(2)}P^{*T}\right)\|_{\text{F}} \nonumber \\
&\leq \|\hat{A}^{(1)} \odot \left(\Xi^{(1)} - P^{*}\Xi^{(2)}P^{*T}\right)\|_{\text{F}} + \| \left(\hat{A}^{(1)} - P^{*}\hat{A}^{(2)}P^{*T} \right)\odot \left(P^{*}\Xi^{(2)}P^{*T}\right)  \|_{\text{F}} \nonumber\\
&\stackrel{(b)}{\leq} \max_{u,v\in[N]} |\hat{A}_{uv}^{(1)}| \| \Xi^{(1)} - P^{*}\Xi^{(2)}P^{*T}\|_{\text{F}} +  \max_{u,v\in[N]} |\Xi_{uv}^{(2)}| \| \hat{A}^{(1)} - P^{*}\hat{A}^{(2)}P^{*T}\|_{\text{F}} \nonumber\\
&\stackrel{(c)}{\leq}  \| \Xi^{(1)} - P^{*}\Xi^{(2)}P^{*T}\|_{\text{F}} + \ell_{\phi}\| \hat{L}^{(1)} - P^{*}\hat{L}^{(2)}P^{*T}\|_{\text{F}}
\label{apd:proofE13}
\end{align}
where (a) is because $P(A\odot B)P^T = (PAP^T) \odot (PBP^T)$  for any $P\in\Pi$ . (b) is because for any two matrices $A, B$, $\|A\odot B\|_{\text{F}}\leq \max_{u,v}|A_{uv}|\|B\|_{\text{F}}$. (c) is because $\max_{u,v\in[N]} |\hat{A}_{uv}^{(1)}|\leq 1, \max_{u,v\in[N]} |\Xi_{uv}^{(2)}|\leq \ell_{\phi}$.
\begin{align} 
\|\hat{A}^{(2)} \odot \Xi^{(2)}\|_{\text{op}} \stackrel{(a)}{\leq} \|\hat{A}^{(2)}\|_{\text{op}} \|\Xi^{(2)}\|_{\text{op}} \stackrel{(b)}{=} \rho(\hat{A}^{(2)})\rho(\Xi^{(2)}) \stackrel{(c)}{\leq}2 \cdot d_{\max}^{(2)}\ell_{\phi} = 2d_{\max}^{(2)}\ell_{\phi}
\label{apd:proofE14}
\end{align}
where (a) is because for any two matrices $A, B$, $\|A\odot B\|_{\text{op}}\leq \|A\|_{\text{op}}\|B\|_{\text{op}}$. (b) is because graphs are undirected and both $\hat{A}^{(2)}$ and $\Xi^{(2)}$ are symmetric matrices. Hence, $\hat{A}^{(2)}$ and $\Xi^{(2)}$ can be diagonalized. For diagonalizable matrices, their operator norms equal their spectral radius $\rho$. (c) is because the following facts: It is known that a degree normalized adjacency matrix $\hat{A}^{(2)}$ has eigenvalues between -1 and 1; And, $\rho(\Xi^{(2)})\leq \max_{v\in[N]} \sum_{u=1}^N|\Xi_{vu}^{(2)}|\leq  2\max_{v\in[N]} d_v^{(2)}\ell_{\phi} = 2d_{\max}^{(2)}\ell_{\phi}$. Here, we use that $\Xi_{vu}^{(2)}$ is not zero iff $vu$ is an edge in the graph.

Lastly, we need to bound $\| \Xi^{(1)} - P^{*}\Xi^{(2)}P^{*T}\|_{\text{F}}$ in Eq.\ref{apd:proofE13}. Let $\pi:[N]\rightarrow[N]$ denote the permutation mapping defined by $P^{*}$, i.e., $P^{*}_{uv}=1$ when $v=\pi(u)$ and $P^{*}_{uv}=0$ otherwise. Pick the $Q\in \text{O}(p)$ that matches the two groups of positional features $Z^{(1)},\,Z^{(2)}$.
\small{\begin{align}
 &\| \Xi^{(1)} - P^{*}\Xi^{(2)}P^{*T}\|_{\text{F}} \nonumber \\
     & \stackrel{(a)}{\leq} \sqrt{\sum_{u,v}\ell_{\phi}^2\left(\|Z_u^{(1)} - Z_v^{(1)}\| - \|Z_{\pi(v)}^{(2)} - Z_{\pi(v)}^{(2)}\|\right)^2}  \nonumber \\
     & = \sqrt{\sum_{u,v}\ell_{\phi}^2\left(\|Z_u^{(1)} - Z_v^{(1)}\| - \|Z_u^{(1)} - Z_{\pi(v)}^{(2)}Q\| + \|Z_u^{(1)} - Z_{\pi(v)}^{(2)}Q\| - \|Z_{\pi(u)}^{(2)}Q - Z_{\pi(v)}^{(2)}Q\| \right)^2}  \nonumber \\
     & \leq \ell_{\phi}\sqrt{2\sum_{u,v}\left(\|Z_u^{(1)} - Z_v^{(1)}\| - \|Z_u^{(1)} - Z_{\pi(v)}^{(2)}Q\|\right)^2 + \left(\|Z_u^{(1)} - Z_{\pi(v)}^{(2)}Q\| - \|Z_{\pi(u)}^{(2)}Q - Z_{\pi(v)}^{(2)}Q\| \right)^2}  \nonumber \\
     & \stackrel{(b)}{\leq} \ell_{\phi}\sqrt{2\sum_{u,v}\left(\|Z_v^{(1)} - Z_{\pi(v)}^{(2)}Q\|^2 + \|Z_u^{(1)} -Z_{\pi(u)}^{(2)}Q\|^2\right)}  \nonumber \\
 & =  2\ell_{\phi}\|Z^{(1)} - P^{*}Z^{(2)}Q\|_{\text{F}} \nonumber \\
 & = 2\ell_{\phi} \eta(Z^{(1)} , P^{*}Z^{(2)}) 
%      &\leq \ell_{\phi} \sqrt{\sum_{u,v}\left|\|Z_u^{(1)} - Z_v^{(1)}\|^2 - \|Z_{\pi(v)}^{(2)} - Z_{\pi(v)}^{(2)}\|^2\right|}  \nonumber \\
%      &\leq \ell_{\phi} \sqrt{\sum_{u,v}\left|\|Z_u^{(1)} - Z_v^{(1)}\|^2 - \|Z_u^{(1)} - Z_v^{(2)}Q\|^2\right| + \left|\|Z_u^{(1)} - Z_v^{(2)}Q\|^2-\|Z_{\pi(v)}^{(2)} - Z_{\pi(v)}^{(2)}\|^2\right|} \nonumber \\
%  &\stackrel{(a)}{\leq}  \ell_{\phi} \|Z^{(1)}Z^{(1)T} - P^{*}Z^{(2)}Z^{(2)T}P^{*T} \|_{\text{F}}  \nonumber \\
%  &\leq \ell_{\phi}\|(Z^{(1)} - P^{*}Z^{(2)}Q)Z^{(1)T} + P^{*}Z^{(2)}Q(Q^TZ^{(2)T}P^{*T} - Z^{(1)T}) \|_{\text{F}}  \nonumber \\
%  &\leq \ell_{\phi}(\|(Z^{(1)} - P^{*}Z^{(2)}Q)Z^{(1)T}\|_{\text{F}} + \|P^{*}Z^{(2)}Q(Q^TZ^{(2)T}P^{*T} - Z^{(1)T}) \|_{\text{F}}) \nonumber \\
%  & \stackrel{(b)}{\leq}  2\ell_{\phi}\|Z^{(1)} - P^{*}Z^{(2)}Q\|_{\text{F}} \nonumber \\
%  & = 2\ell_{\phi} \eta(Z^{(1)} , P^{*}Z^{(2)}) 
 \label{apd:proofE15}
\end{align}}
\normalsize
where (a) is due to the $\ell_\phi$-Lipschitz continuity of the component-wise function $\phi$ and (b) is because of triangle inequalities.%$Z^{(1)}$ and  $P^{*}Z^{(2)}Q$ have orthonormal columns.

Plugging Eq.\ref{apd:proofE15} into Eq.\ref{apd:proofE13} and plugging Eqs.\ref{apd:proofE13},\ref{apd:proofE14} into Eq.\ref{apd:proofE12}, further using Eqs.\ref{apd:proofE10},\ref{apd:proofE11}, we achieve
\begin{align}
    &\|\hat{X}^{(1)}- P^*\hat{X}^{(2)}\|_{\text{F}} \nonumber \\
    \leq &\ell_\psi\ell_{\phi}\|W\|_{\text{op}}\left[(2^{5/2}\delta + 1)\|X^{(1)} \|_{\text{op}}\|L^{(1)}-P^*L^{(2)}P^{*T}\|_\text{F} +2d_{\max}^{(2)}\ell_{\phi}\|X^{(1)}-P^*X^{(2)}\|_{\text{F}} \right]  \nonumber \\
    \leq & (7\delta\|X^{(1)} \|_{\text{op}} + 2d_{\max}^{(2)})\ell_\psi\ell_{\phi}\|W\|_{\text{op}} d(\gG^{(1)},\gG^{(2)}).
    \label{apd:proofE16}
\end{align}
Combining Eq.\ref{apd:proofE16} with the bound on positional features in Eq.\ref{apd:proofE11}, we conclude the proof by 
\begin{align}
    &\|\hat{X}^{(1)}- P^*\hat{X}^{(2)}\|_{\text{F}} + \eta(\hat{Z}^{(1)} , P^{*}\hat{Z}^{(2)}) \nonumber\\
    &\leq [(7\delta\|X^{(1)} \|_{\text{op}} + 2d_{\max}^{(2)})\ell_\psi\ell_{\phi}\|W\|_{\text{op}} + 3\delta]d(\gG^{(1)},\gG^{(2)}).
\end{align}
%Because of the symmetry, we can replace $\|X^{(1)}\|_{\text{op}}$ by $\min\{\|X^{(1)} \|_{\text{op}},\|X^{(2)} \|_{\text{op}}\}$ and $d_{\max}^{(2)}$ by $\min\{d_{\max}^{(1)},d_{\max}^{(2)}\}$. The result keeps true.

\section{Proof of Theorem~\ref{thm:general-pe}}
To prove PE-equivariance, consider two graphs $\gG^{(1)} = (A^{(1)}, X^{(1)})$ and  $\gG^{(2)} = (A^{(2)}, X^{(2)})$ that have perfect matching $P^*$, $L^{(1)} = P^*L^{(2)}P^{*T}$, $X^{(1)} = P^*X^{(2)}$. 

Let $Z^{(i)}$ denote the positional features obtained by decomposing the optimal solution $M^{(i)*}$ to the optimization problem Eq.\ref{eq:pe-unified}. Because $L^{(1)} = P^*L^{(2)}P^{*T}$, we have  $A^{(1)}=P^*A^{(2)}P^{*T}$ and $D^{(1)}=P^*D^{(2)}P^{*T}$. Then,
\begin{align*}
    M^{(1)*} &= \argmin_{M:\text{rank}(M)\leq p} \text{tr}(f_{+}(A^{(1)})g(M) + f_{-}(D^{(1)})g(-M)) \\
    & \stackrel{(a)}{=}  \argmin_{M:\text{rank}(M)\leq p} \text{tr}(P^*f_{+}(A^{(2)})P^{*T}g(M) + P^*f_{-}(D^{(2)})P^{*T}g(-M)) \\
    & \stackrel{(b)}{=}  \argmin_{M:\text{rank}(M)\leq p} \text{tr}(f_{+}(A^{(2)})P^{*T}g(M)P^{*} +f_{-}(D^{(2)})P^{*T}g(-M)P^{*}) \\
    & \stackrel{(c)}{=} \argmin_{M:\text{rank}(M)\leq p} \text{tr}(f_{+}(A^{(2)})g(P^{*T}MP^{*}) +f_{-}(D^{(2)})g(-P^{*T}MP^{*})) \\
    & \stackrel{(d)}{=} P^*M^{(2)*}P^{*T}
\end{align*}
Here (a) is because $A^{(1)}=P^*A^{(2)}P^{*T}$ and the assumptions on $f_+$ and $f_-$. (b) is because for two squared matrices $A,B$, $\text{tr}(AB) = \text{tr}(BA)$. (c) is because $g$ is component-wise function. (d) is because $M^{*(2)}$ is the unique solution of $ \argmin_{M:\text{rank}(M)\leq p} \text{tr}(f_{+}(A^{(2)})g(M) +f_{-}(D^{(2)})g(-M))$.

Recall $M^{(1)*} = Z^{'(1)}Z^{(1)T}$, $Z^{(1)T}Z^{(1)}= I$ and $M^{(2)*} = Z^{'(2)}Z^{(2)T}$, $Z^{(2)T}Z^{(2)}= I$. Note that $Z^{(1)}, Z^{(2)}$ that satisfy such decompositions are not unique. As $\text{rank}(M^{(1)*})=\text{rank}(M^{(2)*})=p$, so $Z^{(1)}$,$Z^{'(1)}$ and $P^*Z^{(2)}$, $P^*Z^{'(2)}$ have full-rank columns.  

Since $M^{(1)*} = P^*M^{(2)*}P^{*T}$, $Z^{'(1)}Z^{(1)T} = P^*Z^{'(2)}Z^{(2)T}P^{*T}$. Because $Z^{'(1)}$ has full-rank columns, $Z^{'(1)T}Z^{'(1)}$ is non-singular. Let $Q = Z^{'(2)T}P^{*T}Z^{'(1)}(Z^{'T(1)}Z^{'(1)})^{-1}$. Then, $Z^{(1)} = P^*Z^{(2)}Q$. Since, $Z^{(1)T}Z^{(1)}=Z^{(2)T}Z^{(2)}=I$, we have $Q^TQ = I$. $Q$ is a squared matrix so $Q\in \text{O}(p)$. That means $Z^{(1)} = P^*Z^{(2)}Q$ for some $Q\in \text{O}(p)$.

%Then, $P^*Z^{(2)}QQ^TZ^{(2)T}P^{*T} =Z^{(1)}Z^{(1)T} = P^*Z^{(2)}Z^{(2)T}P^{*T}$, so $P^*Z^{(2)}(QQ^T - I)Z^{(2)T}P^{*T} = 0$. Use $Z^{(2)T}P^{*T}$ and $P^{*}Z^{(2)}$ to left and right multiple the LHS, and we obtain $Z^{(2)T}Z^{(2)}(QQ^T - I)Z^{(2)T}Z^{(2)} = 0$  Because $Z^{(2)}$ has full-rank columns, $Z^{(2)T}Z^{(2)}$ is non-singular. This implies $QQ^T=I$. $Q$ is a squared matrix so $Q\in \text{O}(p)$. That means $Z^{(1)} = P^*Z^{(2)}Q$ for some $Q\in \text{O}(p)$.

% t use Lemma~\ref{thm:stable}. Because $L^{(1)} = P^*L^{(2)}P^{*T}$ and $\lambda_{p}\neq \lambda_{p+1}$, then there exist $Q\in\text{O}(p)$. %and $b\in\mathbb{R}^p$.
% \begin{align}\label{apd:proofD11}
%     Z^{(1)} = P^*Z^{(2)}Q %+ \mathbf{1}b^T,
% \end{align}
% %where $Z^{(i)}$ is the Laplacian eigenmap of $L^{(i)}$. 

Now, we consider a GNN layer $g$ that satisfies Eqs.~\ref{eq:pi-eq},\ref{eq:oth-eq}. Also denote the output as $(\hat{X}^{(1)}, \hat{Z}^{(1)})=g(A^{(1)}, X^{(1)}, Z^{(1)})$ and $(\hat{X}^{(2)}, \hat{Z}^{(2)}) = g(A^{(2)}, X^{(2)}, Z^{(2)})$.
\begin{align*}
    (\hat{X}^{(1)}, \hat{Z}^{(1)}) & = g(A^{(1)}, X^{(1)}, Z^{(1)})  \\ 
    &\stackrel{(a)}{=} g(P^*A^{(2)}P^{*T}, P^{*T}X^{(2)},  P^*Z^{(2)}Q )\\
    &\stackrel{(b)}{=} P^*g(A^{(2)}, X^{(2)}, Z^{(2)}Q) \\
    &\stackrel{(c)}{=}  (P^*\hat{X}^{(2)}, P^*\hat{Z}^{(2)}Q)
\end{align*}
Here (a) is because the perfect matching between $\gG^{(1)}$ and  $\gG^{(2)}$, and $Z^{(1)} = P^*Z^{(2)}Q$. (b) is due to Eq.~\ref{eq:pi-eq} and (c) is due to Eq.~\ref{eq:oth-eq}.

Therefore, $\hat{X}^{(1)} = P^*\hat{X}^{(2)}$ and $\eta( \hat{Z}^{(1)},  \hat{Z}^{(2)}) = 0$, which implies that $g$ satisfies PE-equivariance.

\section{Review of the E-GNN layer}~\label{apd:EGNN}

\cite{satorras2021n} studies the problem when the nodes of a graph have physical coordinates as features and proposes E-GNN to deal with this kind of graph data. E-GNN aims to keep permutation equivariant with respect to the node order, and translation equivariant, rotation equivariant, reflection equivariant with respect to the physical coordinate features. As E-GNN asks even more (translation equivariance) than Eqs.~\ref{eq:pi-eq},\ref{eq:oth-eq}, E-GNN can be adopted to leverage PE techniques to keep PE-equivariance. The specific form of E-GNN is as follows. 

Given a graph $\gG = (\gV, \gE)$ with nodes $v_i \in \gV$ and edges $e_{ij} \in \gE$. E-GNN layer takes the node embeddings $\textbf{h}^l = \{\textbf{h}_0^l, ..., \textbf{h}_{M-1}^l\}$, coordinate embeddings $\textbf{x}^l = \{\textbf{x}_0^l,...,\textbf{x}_{M-1}^l \}$ and edge information $\gE = (e_{ij})$ as input and outputs $\textbf{h}^{l+1}$ and $\textbf{x}^{l+1}$, respectively. Thus, the E-GNN layer can be denoted as: $\textbf{h}^{l+1}, \textbf{x}^{l+1} = \text{EGCL}(\textbf{h}^{l}, \textbf{x}^{l}, \gE)$. The layer is defined as following:

\begin{align*}
    \textbf{m}_{ij} & =  \phi_e(\textbf{h}_i^{l}, \textbf{h}_j^{l}, ||\textbf{x}_i^l - \textbf{x}_j^l||^2, a_{ij})\\
    \textbf{x}_i^{(l+1)} & = \textbf{x}_i^l + \sum_{j \neq i}(\textbf{x}_i^l - \textbf{x}_j^l)\phi_x(\textbf{m}_{ij})\\
    \textbf{m}_i & = \sum_{j \in \gN_i} (\textbf{m}_{ij})\\
    \textbf{h}_i^{(l+1)} & = \phi_h(\textbf{h}_i^{(l)},\textbf{m}_i)
\end{align*}
 
Where $a_{ij}$ is the edge attributes, $\phi_e$ represents edge operation, $\phi_x$ represents edge embedding operation and $\phi_h$ represents node operation. %The E-GNN layer is translation equivariant, rotation equivariant and permutation equivariant. (Proof could be found in)

\section{The optimization form for general PE techniques} \label{apd:PE-detail} 
Given an undirected and weighted network $\mathcal{G}= (V,E,A)$ with $N$ nodes, LINE \citep{tang2015line} with the second order proximity (aka LINE (2nd)) aims to extract two latent representation matrices $Z, Z' \in \mathbb{R}^{N \times p}$. Let $Z_i, Z_i'$ denote the $i$th row of $Z$ and the $i$th row of $Z'$, respectively. The objective function of LINE (2nd) follows
\begin{align*}
    \max_{Z,Z'}\quad \sum_{i=1}^N\sum_{j=1}^N A_{ij}g(Z_iZ_j'^{T}) + b\sum_{i=1}^N\mathbb{E}_{j'\sim \mathbb{P}_V}(g(-Z_iZ_j'^{T}))
\end{align*}
where $g(x)$ is the log sigmoid function $g(x) = x - \log(1+\exp(x))$ and $b$ is a positive constant. The first term corresponds to the positive examples, i.e., links in the graph while the second term is based on network negative sampling. Also, $\mathbb{P}_V$ is some distribution defined over the node set. LINE adopts $\mathbb{P}_V(j) \propto d_j^{\frac{3}{4}}$ where $d_j$ is the degree of node $j$. By filling the expectation and using matrix form to rewrite the objective, we have 
\begin{align*}
     &\sum_{i=1}^N\sum_{j=1}^N A_{ij}g(Z_iZ_j'^{T}) + b\sum_{i=1}^N\mathbb{E}_{j'\sim \mathbb{P}_V}(g(-Z_iZ_j'^{T})) \\
     =&\sum_{i=1}^N\sum_{j=1}^N A_{ij}g(Z_iZ_j'^{T}) + c\sum_{i=1}^Nd^{\frac{3}{4}}g(-Z_iZ_{j'}'^{T}) \\
     =& \text{tr}(A^Tg(ZZ')) + \text{tr}(cD^{\frac{3}{4}}\mathbf{1}\mathbf{1}^Tg(ZZ')) \\
     =& \text{tr}(Ag(Z'Z)) + c\mathbf{1}\mathbf{1}^TD^{\frac{3}{4}}g(Z'Z))
\end{align*}
where $c = \frac{b}{\sum_{j=1}^N d_j^{3/4}}$. So for LINE, $f_+(A) = A$ and $f_-(D) = c\mathbf{1}\mathbf{1}^TD^{\frac{3}{4}}$. It is easy to validate that for all $P\in \Pi$, $f_+(PAP^T) = Pf_+(A)P^T$ and $f_-(PDP^T) = Pf_-(D)P^T$, which satisfies the condition in Theorem~\ref{thm:general-pe}.

Deepwalk \citep{perozzi2014deepwalk} can be rewritten by following the similar idea. Deepwalk firstly performs random walks with certain length for many times starting from each node and then treat each walk as a sequence of node-id strings. Deepwalk trains a skip-gram model on these node-id strings~\citep{mikolov2013distributed}. Now we consider the skip-gram model with negative sampling (SGNS). Denote the collection of observed words and their context pairs as $\mathcal{D}$. $\#(w,c)$ denotes the number of times that the pair $(w,c)$ appears in $\mathcal{D}$.  $\#(w)$ and $\#(c)$ indicates the number of times $w$ and $c$ occurred in $\mathcal{D}$. Let $Z_w$ denote the vector representation of $w$ and $Z_c'$ denote the vector representation of $c$. According to \citep{levy2014neural}, the objective function of SGNS follows

\begin{align*}
    \sum_{w}\sum_{c} \#(w,c) g(Z_wZ_c') + b\sum_{w}\#(w)\mathbb{E}_{c'\sim \mathbb{P}_c}(g(-Z_wZ_{c'}'^{T}))
\end{align*}
Deepwalk adopts this objective by viewing each node $v$ in $V$ as a word and viewing any node that gets sampled simultaneously with $v$ within $T$ hops  of the random walk as the context $c$. Let $\Phi = D^{-1}A$ denote the random walk matrix of the graph. In this case, for a node $v$ as the word and for another node $u$ as the context, the expected number $\#(v,u) \propto \Phi'_{vu}=\sum_{k=1}^T (d_v(\Phi^k)_{vu}+d_u(\Phi^k)_{uv})$. The expected number $\#(v) \propto d_v$ which uses the stationary distribution of random walk over a connected graph is proportional to the node degrees. We also set the negative context sampling probability $\mathbb{P}_c$ is proportional to the node degrees, i,e., $\mathbb{P}_c(u) \propto d_u$. Then, in Deepwalk, the SGNS objective reduces to 

\begin{align*}
    \sum_{v=1}^N\sum_{u=1}^N \Phi'_{vu} g(Z_vZ_u') + c\sum_{v=1}^N\sum_{u=1}^N d_vd_ug(-Z_vZ_{u}'^{T})
\end{align*}
for some positive constant $c$. Similar to the derivation for LINE, we can rewrite it into the matrix form
\begin{align*}
    \text{tr}(\Phi' g(Z'Z^T) + D\mathbf{1}\mathbf{1}^TDg(-Z'Z^{T})).
\end{align*}
Therefore, for Deepwalk, $f_+(A) = \Phi' = \sum_{k=1}^T (D\Phi^k+\Phi^{Tk}D)$ and $f_-(D) = cD\mathbf{1}\mathbf{1}^TD$,  where $\Phi = D^{-1}A$ and c is a positive constant. It is also easy to verify that  for all $P\in \Pi$, $f_+(PAP^T) = Pf_+(A)P^T$ and $f_-(PDP^T) = Pf_-(D)P^T$, which satisfies the conditions in Theorem~\ref{thm:general-pe}.

As for Node2vec \citep{grover2016node2vec}, it performs a 2nd-order random walk to collect node-id strings and then train an SGNS model. The optimization form is more involved, interested readers could check \citep{qiu2018network,grover2016node2vec} for more details.

\section{Supplement for experiments}\label{apd:exp}

We put more specifics of datasets and baselines adopted in Sec.~\ref{apd:datasets} and Sec.~\ref{apd:baselines}, respectively. We describe how we tune our model in Sec.~\ref{apd:parameter-tuning}. We list further experimental results in Sec.~\ref{apd:analysis}. We further conduct three supplementary experiments t demonstrate the generalization capability, stability and wide applicability of our theory and proposed \proj layer in Secs.~\ref{adp:supp-exp1}-\ref{adp:supp-exp3}.

\subsection{Datasets} \label{apd:datasets}

The citation networks-\textbf{Cora}, \textbf{Citeseer} and \textbf{Pubmed} are collected by \cite{sen2008collective}, where nodes  represent documents and edges (undirected) represent citations. Node features are the bag-of-words representation of documents. The Cora dataset contains 2708 nodes, 5429 edges and 1433 features per node. The Citeseer dataset contains 3327 nodes, 4732 edges and 3703 features per node. The Pubmed dataset contains 19717 nodes, 44338 edges and 500 features per node.

\textbf{Twitch} is obtained from \cite{rozemberczki2021multi}. Twitch is a user-user networks of gamers, where nodes correspond to users and edges correspond to mutual friendship between them. Node features correspond to the games they played and liked, users' location and streaming habits. Twitch contains 7 user networks over different countries, including Germany (DE), England (EN), Spain (ES), France (FR), Portugal (PT) and Russia (RU). The details of these datasets are shown in Table~\ref{table4}.

\textbf{Chameleon} is used in \cite{chien2021adaptive}. Chameleon is a page-page network on topic `Chameleon' in Wikipedia (December 2018), where nodes correspond to articles and edges represent mutual links between the articles. Node features indicate the presence of several informative nouns in the articles and the average monthly traffic (October 2017 - November 2018). Chameleon dataset contains 2277 nodes, 36101 edges and 2325 features per node.

Protein-protein interaction (PPI) dataset contains 24 graphs corresponding to different human tissues \citep{zitnik2017predicting}. We adopt the preprocessed data provided by \cite{hamilton2017inductive} to construct graphs. The average number of nodes per graph is 2372 and each node has 50 features, which correspond to positional gene sets, motif gene sets and immunological signatures.

Ogbl-ddi and ogbl-collab are chose from the open graph benchmark (OGB) \citep{hu2021ogb}, which adopt more realistic train/validation/test splitting, such as by time (ogbl-collab) and by by drug target in the body (ogbl-ddi). ogbl-ddi is a drug-drug interaction network, where nodes represent drugs and edges represent interaction between drugs, where the joint effect of taking the two drugs together is considerably different from the effect that taking either of them independently. Ogbl-collab is an author collaboration graph, where nodes represent authors and edges represent the collaboration between authors. The 128 dimension node features of ogbl-collab is extracted by by averaging the word embeddings of the authors' papers. Ogbl-ddi has 4267 nodes and 1.3M edges. Obdl-collab has 0.23M nodes and 1.3M edges.

\begin{table}[t]
\caption{Summary of Twitch dataset}
\label{table4}
\centering
\begin{tabular}{c|c|c|c|c|c|c}
\hline
         & DE      & EN     & ES     & FR      & PT     & RU     \\ \hline
Nodes    & 9,498   & 7,126  & 4,648  & 6,549   & 1,912  & 4,385  \\
Edges    & 153,138 & 35,324 & 59,382 & 112,666 & 31,299 & 37,304 \\
Features & 3,170   & 3,170  & 3,170  & 3,170   & 3,170  & 3,170  \\ \hline
\end{tabular}
\end{table}

\subsection{Baseline details} \label{apd:baselines}

We have 4 baselines based on GNNs and 2 baselines based on network embedding techniques, namely, LE \citep{belkin2003laplacian} and Deepwalk (DW) \citep{perozzi2014deepwalk}. We will first introduce the implementation of LE and DW, then discuss other baselines.

Both LE and DW embed the networks in $\mathbb{R}^{128}$ in an unsupervised way. For LE, we factorize of the graph Laplacian matrix: $\Delta = I - D^{-\frac{1}{2}}AD^{-\frac{1}{2}} = U \Lambda U^T $, where $A$ is the adjacency matrix, $D$ is the degree matrix, and $\Lambda$, $U$ correspond respectively to the eigenvalues and eigenvectors. We use the 128 smallest eigenvectors as LE. For DW, we use the code provide by OpenNE\footnote{https://github.com/thunlp/OpenNE/tree/pytorch}. Both methods use the inner product between pairwise node representations as the link representations. Then, the link representations are fed to an MLP for final predictions.

Regarding to GNN-based baselines,  VGAE is implemented according to the code\footnote{https://github.com/tkipf/gae} \citep{kipf2016variational} given by the original paper, with 2 message passing layers
with 32 hidden dimensions. P-GNN is implemented by adopting the code\footnote{https://github.com/JiaxuanYou/P-GNN} provided by the original paper \citep{you2019position}, with 2 message passing layers with 32 dimensions, with a tuned dropout ratio in \{0, 0.5\}. GNN transformer layer is implemented by adopting the code\footnote{https://github.com/graphdeeplearning/graphtransformer} provided by the original paper \citep{dwivedi2020generalization} with 2 message passing layers with 128 hidden dimensions. SEAL is implemented by adapting the code\footnote{https://github.com/muhanzhang/SEAL} provided by the original paper. Notice the ogbl-ddi graph contains no node features, so the we use free-parameter node embeddings as the input node features and train them together with the GNN parameters. We slightly tune the hidden dimensions and layers of these baselines and present the best results.

\begin{figure}[t]
\centering
%\vspace{-1cm}
    \includegraphics[trim={2.9cm 0.3cm 4.0cm 0.9cm},clip,width=0.49\textwidth]{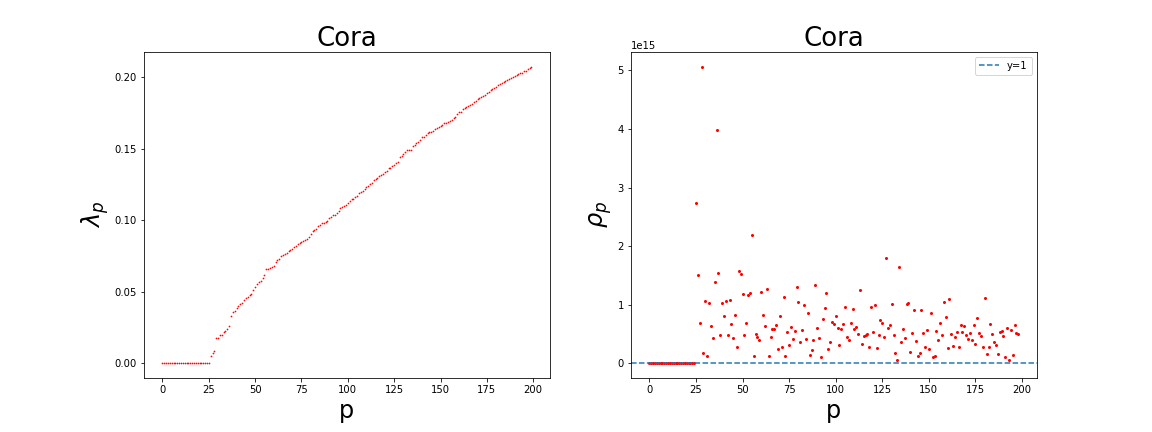}
	\includegraphics[trim={2.9cm 0.3cm 4.0cm 0.9cm},clip,width=0.49\textwidth]{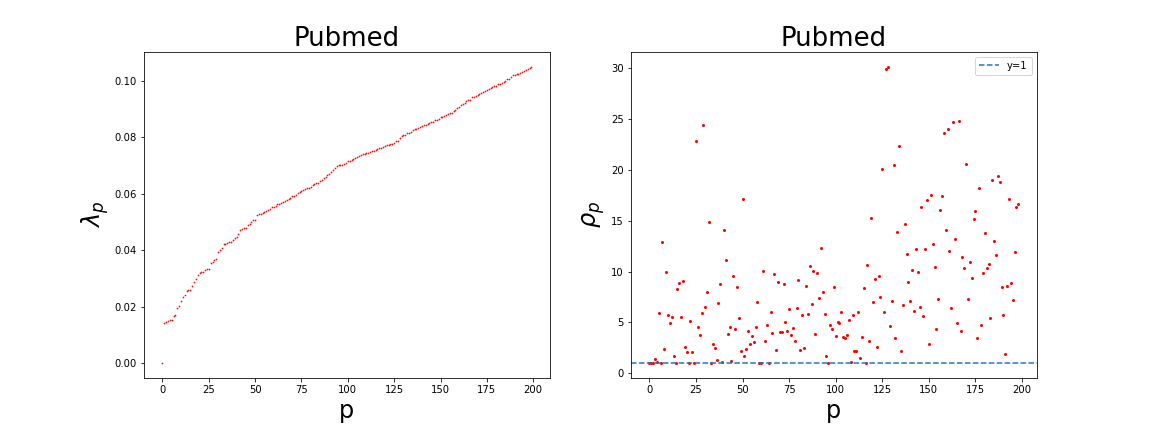}
	\includegraphics[trim={2.9cm 0.3cm 4.0cm 0.9cm},clip,width=0.49\textwidth]{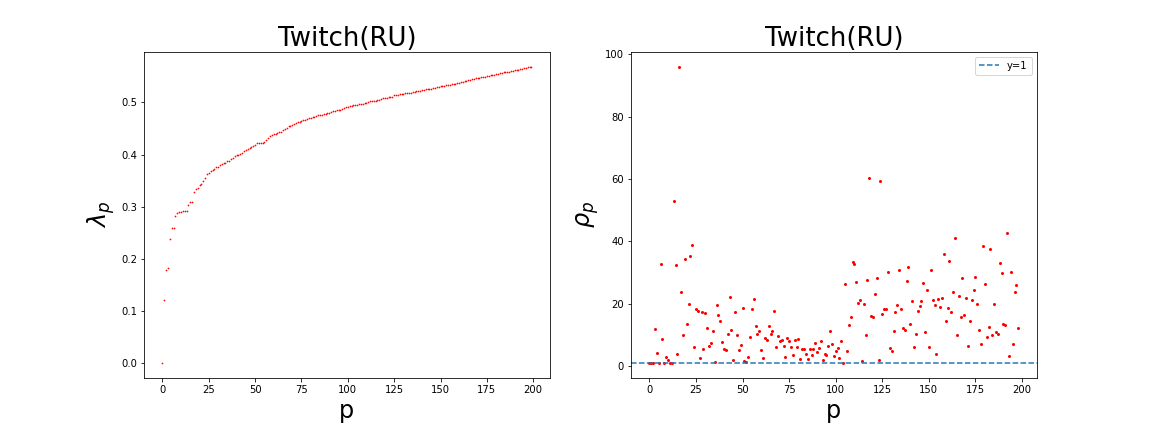}
	\includegraphics[trim={2.9cm 0.3cm 4.0cm 0.9cm},clip,width=0.49\textwidth]{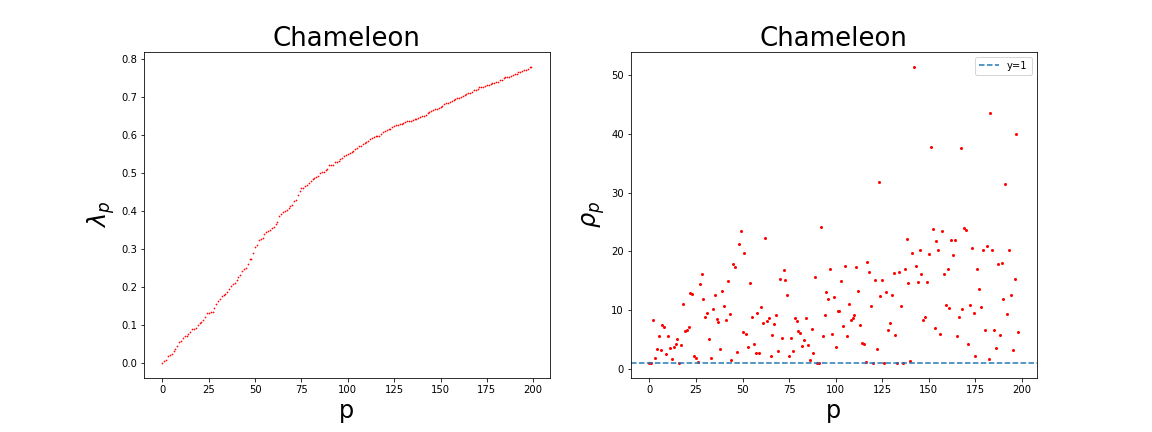}
    %\vspace{-0.1cm}
    \caption{\small{Eigenvalues $\lambda_p$ and the stability ratio $\rho_p=\frac{|\lambda_p - \lambda_{p+1}|}{\min_{1\leq k\leq p}|\lambda_k - \lambda_{k+1}|}$. \proj with sensitivity $|\lambda_p - \lambda_{p+1}|^{-1}$  is far more stable than previous methods with sensitivity $\max_{1\leq k\leq p}|\lambda_k - \lambda_{k+1}|^{-1}$. Over Cora, $\rho_p$ is extremely large because there are multiple eigenvalues ($\rho_p$ is still finite due to the numerical approximation of the eigenvalues). Over Pubmed, Twitch(RU) and Chameleon even if there are no multiple eigenvalues, $\rho_p$ is mostly larger than 5.}}
     %\vspace{-0.3cm}
    \label{fig:otherEigen}
\end{figure}

\begin{figure}[t]
\centering
    \includegraphics[width=0.33\textwidth]{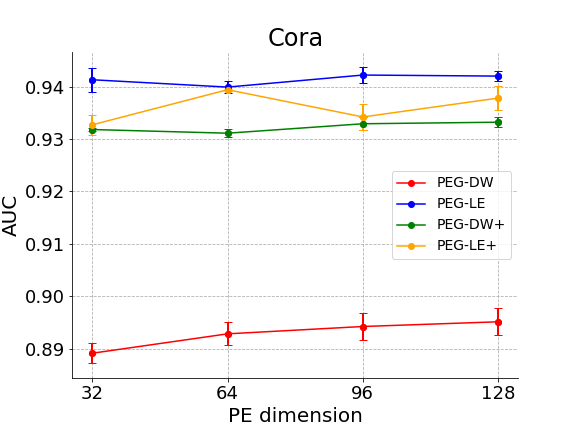}
	\includegraphics[width=0.335\textwidth]{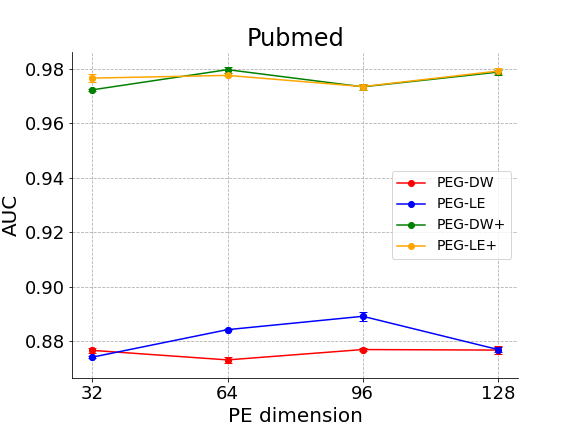}
	\includegraphics[width=0.32\textwidth]{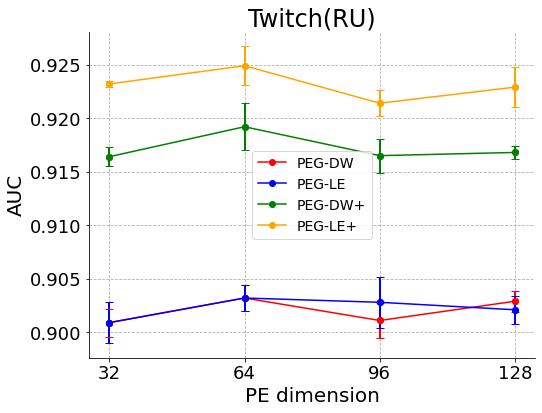}
    %%\vspace{-0.6cm}
    \caption{\small{The performance (AUC) of \proj with different dimensions of positional encodings for traditional link prediction (Task 1).}}
    \label{fig:dim}
\end{figure}

Moreover, we aim to understand whether \proj is sensitive to the dimension of positional features. We conduct experiments on Cora, Pubmed and Twitch(RU) to test the sensitivity of the dimension positional features. In Fig.~\ref{fig:dim}, we compare the performance of \proj-DW, \proj-LE, \proj-DW+ and \proj-LE+ with PE in different dimensions. All the experiment settings follow the setting to get Table~\ref{table1} except for the dimension of positional features.

\subsection{Hyperparameters Tuning for PEG} \label{apd:parameter-tuning}
Table~\ref{table5} lists the most important hyperparameters, which applies to our proposed model PEG. Grid research is used to find the best hyperparameter combination. Note that our model actually got very slightly tuned. We believe more extensively hyperparameter tuning may yield even better results. The models are trained until the loss converges and we report the best model by running each for 10 times. We provide our code for reproducing the experimental results to the reviewers and area chairs in the discussion forums.

\begin{table}[H]
\caption{List of hyperparameters and their value/range}
\label{table5}
\centering
\begin{tabular}{cc}
\hline
\textbf{Hyperparameters} & \textbf{Value/Range} \\ \hline
batch size               & 64, 128, 64$\times$1024(ogb)              \\
learning rate            & 1e-2, 1e-3           \\
optimizer                & Adam                 \\
conv. layers             & 2                    \\
conv, hidden dim         & 128                  \\
PE. dim                  & 128                  \\ \hline
\end{tabular}
\end{table}

\subsection{Further analysis} \label{apd:analysis}

As a supplement to the Lemma~\ref{thm:unstable} and Lemma~\ref{thm:stable}, we compute the ratio between the inverse eigengap between the $p$th and $(p+1)$th eigenvalues $\{|\lambda_p - \lambda_{p+1}|^{-1}\}$ and $\max_{1\leq k\leq p} |\lambda_k - \lambda_{k+1}|^{-1}$ over more graphs as shown in Fig.~\ref{fig:otherEigen}.

Fig.~\ref{fig:dim} shows that different PE techniques have different dimensional sensitivity. In general, DW seems to be more stable than LE over the three datasets and \proj-DW+ achieves stable performance. When the dimension increases, \proj are more likely to achieve higher performance, but will be more time consuming.

\subsection{Edge weight visualization}\label{adp:edge_weight}

\begin{figure}[h]
\centering
%\vspace{-1cm}
    %\includegraphics[trim={0.3cm 0.1cm 1.4cm 0.7cm},clip,width=0.32\textwidth]{FIG/Cora_edge_weight.png}
	%\includegraphics[trim={0.3cm 0.1cm 1.4cm 0.7cm},clip,width=0.32\textwidth]{FIG/Citeseer_edge_weight.png}
	%\includegraphics[trim={0.3cm 0.1cm 1.4cm 0.7cm},clip,width=0.32\textwidth]{FIG/Pubmed_edge_weight.png}
	\includegraphics[trim={0.3cm 0.1cm 1.4cm 0.7cm},clip,width=0.32\textwidth]{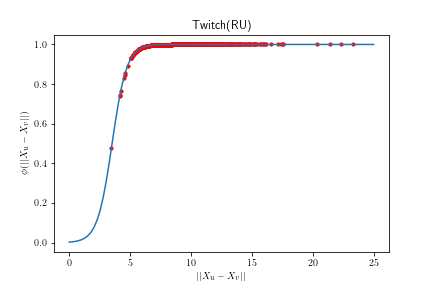}
	\includegraphics[trim={0.3cm 0.1cm 1.4cm 0.7cm},clip,width=0.32\textwidth]{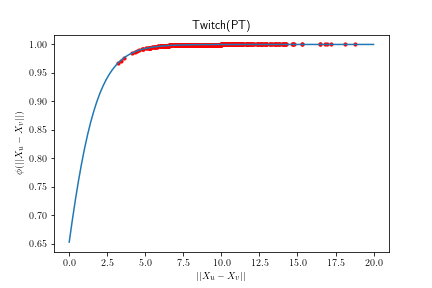}
	\includegraphics[trim={0.3cm 0.1cm 1.4cm 0.7cm},clip,width=0.32\textwidth]{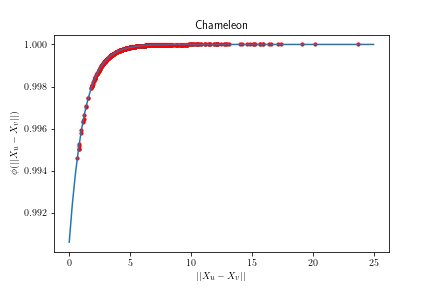}
    %\vspace{-0.1cm}
    \caption{The edge weight transformation curve of PEG for each dataset. 500 selected edges are represented as red points.}
     %\vspace{-0.3cm}
    \label{fig:edge-weight}
\end{figure}

In the PEG layer, we calculate an edge weight according to the distance between the representations of the end nodes of the edge. To further understand the learnt relationship between edge weights and the distance between node representations, we visualize the edge weight transformation curves. We run the experiment traditional link prediction (task 1) and use DW as the node positional embedding, i.e., the model \proj-DW. For each dataset, we draw the edge weight transformation curves in Fig.~\ref{fig:edge-weight}. Note that each curve ranges in x-axis from the smallest distance to the largest distance observed from the corresponding graph, we also randomly select 500 distances from each graph and scatter them on the curve as shown in Fig.~\ref{fig:edge-weight}. Fig.~\ref{fig:edge-weight} shows that the edge weight increases monotonically with respect to the input distance, but the relationship between them is non-linear. Most of the weights are close to 1 but a few weights are much smaller. %The curve demonstrates that the edges between nodes far away from each other are more important. 

\subsection{Link prediction over Random graphs}\label{adp:supp-exp1}

\begin{figure}[t]
\centering
    \includegraphics[width=0.5\textwidth]{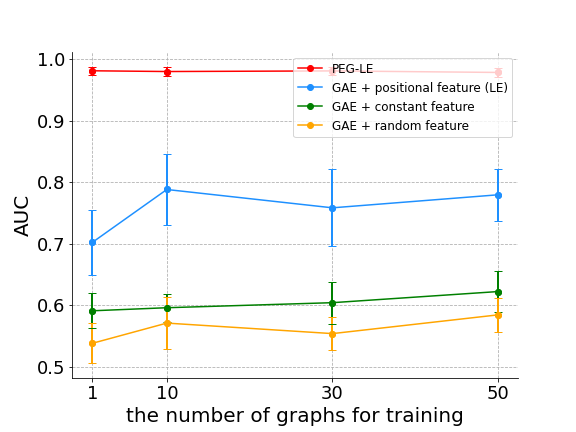}
    %\vspace{-0.6cm}
    \caption{\small{The comparison of AUC-ROC between \proj and GAE baselines for link prediction on random graphs.}}
    \label{fig:training_size}
\end{figure}

To further demonstrate the generalization of our model, we conduct inductive link prediction over random graph models. Specifically, we use stochastic block models (SBM) with two blocks~\cite{holland1983stochastic} to generate random graphs. Each block contains 500 nodes and the probability to form a link between two nodes within each block is 0.3, while the probability to form a link between two nodes across different blocks is 0.1. Here, we randomly select links inside the blocks as positive samples, and randomly select the missing links (unconnected node pairs) from the generated graph as negative samples. The model is trained by using various numbers of graphs (e.g. 1, 10, 30, 50), validated and tested on 10 random graphs respectively. For each graph that is used for training or testing, we utilize 10\% positive links and pair them with the same number of negative missing links for training or testing respectively. The rest settings are remained the same as in Task 2. We choose three variants of GAE that use constant features (node degree), random features and positional features as baselines. All the models are trained until they converge and the models with the best validation performance are used to report the results (averaged by 10 independent runs), which are shown in Fig.~\ref{fig:training_size}.

For various sizes of input graphs, \proj achieves AUC around 0.98 by just using one training graph and slightly improves AUC by using more training graphs, which consistently outperforms all three GAE variants that even use the entire 50 training graphs with large margins. This directly demonstrates the stability and better generalization of \proj models.

\subsection{The study of PE stability by perturbing the input graphs} \label{adp:supp-exp2}
\begin{figure}[t]
\centering
%\vspace{-1cm}
    \includegraphics[width=0.45\textwidth]{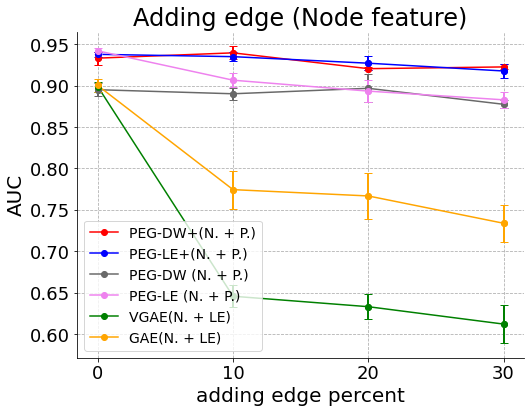}
	\includegraphics[width=0.45\textwidth]{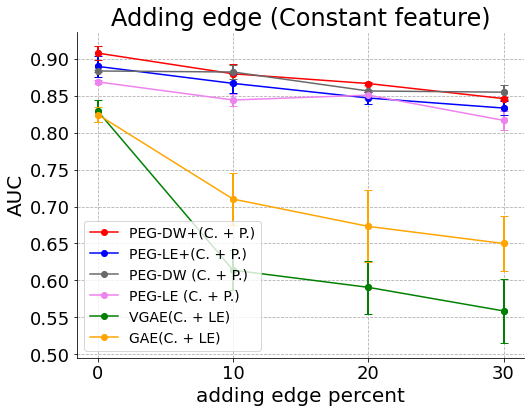}
	\includegraphics[width=0.45\textwidth]{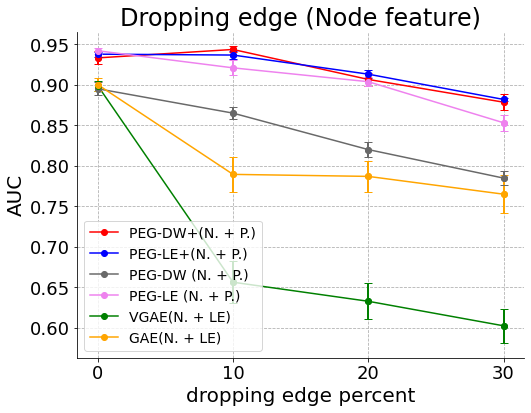}
	\includegraphics[width=0.45\textwidth]{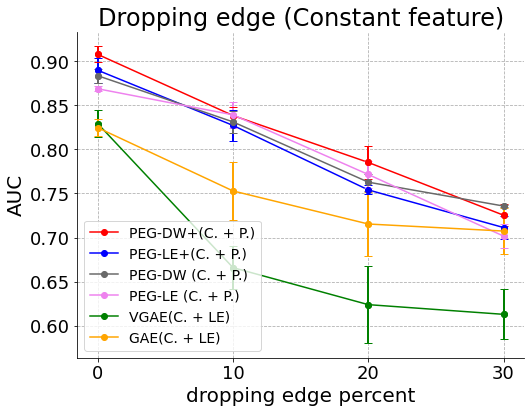}
    %\vspace{-0.1cm}
    \caption{The AUC-ROC of \proj and GAE/VGAE for link prediction in terms of PE stability (w. node or constant feature and w. adding or dropping edges). To perturb the input graphs, we can randomly add or delete some edges with the percentage shown in the X-axis before the inference.}
     %\vspace{-0.3cm}
    \label{fig:perturbation}
\end{figure}

To further investigate PE stability, we consider two versions of models utilizing PE -- one is \proj, which is PE equivariant and stable, and the others are GAE/VGAE (N./C. + P.), which fail to satisfy the two equivariant properties and thus may not be stable. These two versions of models are evaluated on the Cora dataset. All the models are trained over the original graph. Then, we perturb the input graph by randomly adding or dropping a certain percentage of links during the inference. %The links added are randomly sampled from the set of missing links that exclude the missing links used for validation and testing. The links dropped are randomly sampled from the set of links that exclude the links used for validation and testing. 
The rest of the settings remain the same for link prediction. Specifically, after training the model, we inject a perturbation on top of the input graph by adding extra links (negative samples) or dropping positive links with the ratio of 10\%, 20\% and 30\%, respectively. We recompute the positional encodings to reflect the structural perturbation of the given graph and plug in the perturbed positional encodings into different models. Other experiment settings remain the same as in Task 1. The results are summarized and plotted in Fig.~\ref{fig:perturbation}.

 By comparing the models with different levels of input perturbation, we can observe from Fig.~\ref{fig:perturbation} that \proj achieves consistently and significantly better performance with either node features or constant features in all four scenarios, which is far more robust than GAE/VGAE-based models that adopt PE under the same level of perturbations. Compared with adding edges, removing edges is more destructive for small-size graphs such as Cora, and thereby does more harm to the model performance. Particularly, all the models using C. + P. achieve subpar performance when considerable number of edges are removed. However, \proj can still show strong resilience against this type of perturbation (especially with node features), while GAE/VGAE generally collapses. These results further support that the performance boost indeed benefits from the equivariance and stability with PEs instead of simply using PE variants.

\subsection{Preliminary results for node classification based on PE}
\label{adp:supp-exp3}
% \htyin{add how to modify \proj to adopt node classification tasks}

In order to evaluate \proj on a wider category of tasks, we consider node classification on citation networks -- \textbf{Cora}, \textbf{Citeseer} and \textbf{Pubmed} \citep{sen2008collective} and closely follow the transductive experimental setup of \cite{yang2016revisiting}. Suppose the final readout of \proj is denoted as $(\hat{X},\,Z)$. To classify the node $u$, we only utilize its node representation $\hat{X}_u$ to make the final predictions. The rest parts and the hyperparameters of the model are kept the same as the one that is good for our task 1 (traditional link prediction). We did not specifically tune the model for node classification. The results of evaluation on node-level are summarized in Table~\ref{table6}. We report the mean accuracy with standard deviation of classification on the test set of our method after 10 runs. 

\begin{table}[h]
\small
\caption{Summary of results in terms of node classification accuracy (mean ± std\%), for Cora, Citeseer and Pubmed.}
\label{table6}
\centering
\begin{tabular}{c|c|c|c}
\hline
\textbf{Method} & \textbf{Cora} & \textbf{Citeseer} & \textbf{PubMed} \\ \hline
\proj-DW          & 82.24 ± 0.02        & 71.22 ± 0.01      & 79.83 ± 0.02    \\
\proj-LE          & 82.16 ± 0.02        & 71.93 ± 0.01      & 79.85 ± 0.01    \\ \hline
GCN             & 81.50 ± 0.05        & 70.38 ± 0.05      & 79.03 ± 0.03    \\
GAT             & 83.03 ± 0.07        & 72.52 ± 0.07      & 79.06 ± 0.03    \\ \hline
\end{tabular}
\end{table}

 Both \proj-LE and \proj-DW significantly outperform GCN and provide comparable results against GAT, which illustrates the effectiveness of the proposed \proj on node level tasks. Recall that we almost did not tune our model to achieve such results. Even better results may be expected by further finer tuning the model on each dataset. Moreover, GAT uses multi-head attention to aggregate information from neighbours where the attention weights are based on node features, while \proj only adopts GCN layers with some edge weights derived from positional encodings that merely depend on the network structures. \proj and GAT essentially utilize orthogonal information source to re-weight the edges. Hence, \proj should be easily further combined with the attention mechanism based on node features to get even better predictions, which this is out of the scope of this work, so we leave it for future study.

\hide{
GAE is sensitive to perturbation and achieve subpar performance, while \proj can achieve better performance and is more stable than GAE/VGAE where there is a perturbation, which demonstrates that using PE in an equivariant and stable way can make the model robust. By comparing our model with GAE that uses constant features and positional encodings, we demonstrate the importance of keeping PE stable. Interestingly, all the models using C. + P. achieve subpar performance when cutting edges before testing. It might because the Cora graph is small and cutting edges hurt the graph structure seriously.}

\end{document}